%% file: paper.tex
\documentclass[twoside]{article}
\input{Definitions}

\usepackage[accepted]{aistats2025}
\usepackage[round]{natbib}

\usepackage[inline,shortlabels]{enumitem}
\usepackage{adjustbox}
\usepackage{microtype}
\usepackage[hyphens]{url}
\usepackage{graphicx}
\usepackage[pdfencoding=auto,psdextra,colorlinks,linkcolor=red,citecolor=blue,urlcolor=blue]{hyperref}
\usepackage{pdfpages}
\usepackage{subcaption}
\usepackage{booktabs}
\usepackage{tikz}
\usepackage{circledsteps}
\newcommand{\tpr}{\text{TPR}}
\graphicspath{{figures/}}

\newcommand{\highlight}[1]{\begingroup\color{black}#1\endgroup}
\begin{document}

%

%

\twocolumn[

\aistatstitle{Fairness Risks for Group-Conditionally Missing Demographics}

\aistatsauthor{ Kaiqi Jiang$^1$ \And Wenzhe Fan$^1$ \And  Mao Li$^2$ \And Xinhua Zhang$^1$ }


\aistatsaddress{ $^1$ Department of Computer Science, University of Illinois at Chicago, Chicago IL 60607 \\
$^2$ Amazon} ]

\begin{abstract}
Fairness-aware classification models have gained increasing attention in recent years as concerns grow on discrimination against some demographic groups. Most existing models require full knowledge of the sensitive features, which can be impractical due to privacy, legal issues, and an individual's fear of discrimination. The key challenge we will address is the group dependency of the unavailability, e.g., people of some race may be more reluctant to reveal their race. Our solution augments general fairness risks with probabilistic imputations of the sensitive features, while jointly learning the \textit{group-conditionally} missing probabilities in a variational auto-encoder. Our model is demonstrated effective on both image and tabular datasets, 
achieving an improved balance between accuracy and fairness.
\end{abstract}

\section{Introduction}
As machine learning systems rapidly acquire new capabilities and get widely deployed to make human-impacting decisions, 
ethical concerns such as fairness have recently attracted significant effort in the community.
To combat societal bias and discrimination in the model that largely inherit from the training data,
a bulk of research efforts have been devoted to addressing group unfairness, 
where the model performs more favorably (\eg, accurately) to one demographic group than another \citep{hardt2016equality,dwork2012fairness,ye2020unbiased,buolamwini2018gender,gianfrancesco2018potential,mehrabi2021survey,yapo2018ethical}.  
Group fairness typically concerns both the conventional classification labels (\eg, recidivism) and socially sensitive group features (\eg, gender, or race).
As applications generally differ in their context of fairness,
a number of quantitative metrics have been developed such as demographic parity, equal opportunity, and equalized odds.
A necessarily outdated overview is available at \citet{barocas2019fairness} and \citet{wikipedia2023fairness}.

Learning algorithms for group fairness can be broadly categorized into pre-, post-, and in-processing methods. 
Pre-processing methods transform the input data to remove dependence between the class and demographics according to a predefined fairness constraint \citep{kamiran2012data}.
Post-processing methods warp the class labels (or their distributions) from any classifier to fulfill the desired fairness criteria \citep{hardt2016equality}.
In-processing approaches, which are most commonly employed including this work, 
learn to minimize the prediction loss while upholding fairness regularizations simultaneously.
More discussions are given in Section~\ref{sec:related}.

Most of these methods require accessing the sensitive demographic feature,
which is often unavailable due to fear of discrimination and social desirability \citep{krumpal2013determinants},
privacy concerns, and legal regulations \citep{coston2019fair,lahoti2020fairness}.
For example, people of some gender may be more inclined to withhold their gender information when applying for jobs dominated by other genders.
Recently several methods have been developed for this data regime.
\citet{lahoti2020fairness} achieved Rawlsian max-min fairness by leveraging computationally-identifiable errors in adversarial reweighted learning.
\citet{hashimoto2018Fairness} proposed a distributionally robust optimization to minimize the risk over the worst-case  group distribution.
However, they are not effective in group fairness and cannot be customized for different group fairness metrics.
\citet{yan2020fair} infers groups by clustering the data, 
but there is no guarantee that the uncovered groups are consistent with the real sensitive features of interest.
For example, the former may identify race while we seek to be fair in gender.

We aim to tackle these issues in a complementary setting,
where demographics are \textit{partially} missing.
Although their availability is limited,
it is still often feasible to obtain a \textit{small} amount of labeled demographics.%
\footnote{According to \citet{Consumer23}, ``A creditor shall not inquire about the race, color, religion, national origin, or sex of an applicant.'' 
Exceptions include inquiry ``for the purpose of conducting a self-test that meets the requirements of §1002.15'', which essentially tests if the fair lending rules are conformed with.}
For example, some people may not mind disclosing their race or gender.
Therefore, different from the aforementioned works that assume complete unavailability of demographics,
we study in this paper the \textit{semi-supervised} setting where they can be partially available at random in both training and test data.
Similarly, the class labels can be missing in training.

Semi-supervised learning (SSL) has been well studied,
and can be applied to impute missing demographics.
\citet{ZhaLon21} simply used the labeled subset to estimate the fairness metrics,
but their focus is on analyzing its estimation bias instead of using it to train a classifier.
\citet{jung2022learning} assigned pseudo group labels by training an auxiliary group classifier, and assigned low-confidence samples to random groups.
However, the confidence threshold needs to be tuned based on the conditional probability of groups given the class label,
which becomes challenging as the latter is also only partially available in our setting.

\citet{DaiWan21} used SSL on graph neural networks to infer the missing demographics, 
but rounded the predictions to binary group memberships.
This is suboptimal because, due to the scarcity of labeled data,
there is  marked \textit{uncertainty} in demographic imputations.
As a result, the common practice of ``rounding'' sensitive estimates---%
so that fairness metrics defined on categorical memberships can be enforced instead of mere independence---%
may over-commit to a group just by chance,
dropping important uncertainty information when the results are fed to downstream learners.
Moreover, most applications do not naturally employ an underlying graph,
and the method only enforces min-max (i.e., adversarial) fairness instead of  group fairness metrics.

Our \textbf{first} contribution, therefore, 
is to leverage the \textit{probabilistic} imputation by designing a differentiable fairness risk in Section \ref{sec:risk_model},
such that it is customized for a \textit{user-specified} fairness metric with discrete demographics as opposed to simple independence relationships,
and can be directly integrated into general SSL algorithms to regularize the learned posterior towards low risks in both classification and fairness.

Our method is \textbf{distinguished} from generic SSL recipes which predict unobserved protected attributes and then optimize  the fairness risk.
Performing the two steps separately prevents the common backbone features to be synthesized that concurrently facilitate identifying demographics and class labels.
This could be ameliorated via a bi-level optimization,
but its computational cost can be high.
We hence resort to flattening the two levels into one joint optimization,
and then address the pathological phenomenon where the group memberships are also ``learned'' to promote fairness.

Interestingly, we found an efficient solution by stopping the gradient of fairness risk with respect to the group estimates.
Furthermore, to effectively implement this principle,
we employed a Monte-Carlo evaluation of the risk,
which significantly accelerates inference and differentiation via a provable $O(1/\epsilon^2)$ sample complexity.
So our \textbf{second} contribution
hits two birds with one stone, 
as detailed in Section~\ref{sec:monte_carlo}.


As our \textbf{third} contribution,
we instantiated the semi-supervised classifier with a new encoder and decoder in a variational autoencoder \citep[VAE,][]{Narayanaswamy2017learning,Kingma2014semi},
allowing \textbf{group-conditional missing demographics},
which has so far only been considered for noisy but not missing demographics.
In general, the chance of unavailability does depend on specific groups -- if people of a certain race are aware of the unfairness against them,
they will be more reluctant to disclose their race.
The comprehensive model will be introduced in Section~\ref{sec:risk_model_vae}.

Our \textbf{fourth} contribution is to demonstrate, in Section~\ref{sec:experiment}, 
that our method empirically outperforms state of the art for fair classification where both the demographics and class labels are only partially available.
The paper is primed with preliminaries  in Section~\ref{sec:prelim}.

\vspace{-0.25em}
\section{Related Work}
\label{sec:related}
\vspace{-0.25em}

Noisy demographics have been tackled by, \eg, \citet{lamy2019noise,celis2021fair}.
Adversarially perturbed and privatized demographics were studied by \citet{celis2021fairb} and \citet{mozannar2020fair},
which also account for group-conditional noise.
However, although they conceptually subsume unavailability as a type of noise,
their method and analysis do not carry through.
Some of these methods, along with \citet{wang2020robust}, 
also require an auxiliary dataset to infer the noise model,
which is much harder for missing demographics.
\citet{shah2023group} dispenses with such a dataset,
but its favorable theoretical properties apply only to Gaussian data,
while the bootstrapping-based extension to non-Gaussian data does not model the conditional distribution of demographics.

Another strategy to tackle missing demographics resorts to proxy features \citep{gupta2018proxy,chen2019fairness,kallus2022assessing}.
Based on domain knowledge, 
they are assumed to correlate with and allude to the sensitive feature in question.
\citet{ZhaDaiShuWan22} optimizes the correlations between these features and the prediction,
but it is not tailored to the specific group fairness metric in question.
This is difficult because it relies on the binary group memberships which is not modeled by the method.
\citet{Zhuetal23} considered the accuracy of estimating the fairness metrics using proxies,
but did not demonstrate its effectiveness when applied to train a fair classifier.
In addition, proxy features can often be difficult to identify.
In face recognition where the sensitive feature is age, race, or wearing glasses, 
raw pixels are no good proxies and identifying semantic features as proxies can be challenging.

\vspace{-0.15em}
\section{Preliminary}
\label{sec:prelim}
\vspace{-0.15em}

We first consider supervised learning with group fairness,
where samples are drawn from a distribution $D$ over $\Xcal \times \{0,1\} \times \{0,1\}$, 
with $(X, A, Y) \sim D$.
Here $X$ is the non-protected feature,
$Y$ is the binary label,
and $A$ is the binary sensitive feature which is also referred to as the group demographics.
Our method can be extended to \textbf{multi-class} labels and sensitive features,
under any fairness risk defined for fully observed $A$ and $Y$.
The details are deferred to Appendix~\ref{sec:multiclass_fairness}.
Since our focus is on addressing missing demographics,
we will stick with the binary formulation for ease of exposition.

We follow the setup in \citet{menon2018cost} and aim to find a measurable \textit{randomized classifier} parameterized by a function $f: \Xcal \to [0,1]$ 
(\eg, a neural network),
such that it predicts an $X \in \Xcal$ to be positive with probability $f(X)$.
That is, the predicted label $\Yhat \in \{0, 1\}$ follows the Bernoulli distribution $\Yhat | X \sim \text{Bern}(f(X))$.
We denote the conditional distribution as $P_f(\Yhat|X)$,
and omit the subscript $f$ for brevity when the context is clear.
This setup also subsumes a classifier based on a \textit{class-probability estimator},
where $f(X) \in [0,1]$ is translated to a hard deterministic label via a learnable threshold $\tau$:
%
$P_f(\Yhat = 1 | X) = \sembrack{f(X) > \tau}$.
Here, the Iverson bracket $\sembrack{\cdot} = 1$ if $\cdot$ is true, 
and 0 otherwise.

Suppose we have a training set $\Dcal_{tr} := \{\xvec_i, y_i, a_i\}_{i=1}^n$ with size $n$,
where $\xvec_i \in \Xcal$ is the nonsensitive feature of the $i$-th example,
$y_i$ is its label,
and $a_i$ is its sensitive feature.
In a completely observed scenario, 
both $y_i$ and $a_i$ are $0$ or $1$,
and we will later extend it to missing values.
As a shorthand, let $\avec = (a_1, \ldots, a_n)^\top$
and $\yvec = (y_1, \ldots, y_n)^\top$.
%
%
We first review some standard group fairness metrics:

\noindent Demographic parity \citep{dwork2012fairness}:
\begin{equation}
\label{eq:def_dem_par}
     P(\Yhat = 1 | A = 0) = P(\Yhat = 1 | A = 1),
\end{equation}
Equalized odds \citep{hardt2016equality}:
for $y \! \in \! \{0, \!1\}$
\begin{equation}
\begin{split}
\label{eq:def_eq_odds}
    P(\Yhat = 1 | A = 0, Y = y) = \!
    P(\Yhat = 1 | A = 1, Y = y),\!
\end{split} 
\!
\end{equation}
Equal opportunity \citep{hardt2016equality}:
\begin{equation}
\label{eq:def_eq_opp}
    P(\Yhat = 1 | A = 0, Y = 1) = \!P(\Yhat = 1 | A = 1, Y = 1). \!\!
\end{equation}

In general, enforcing perfect fairness can be too restrictive,
and approximate fairness can be considered via fairness measures.
For example, the mean difference compares their difference \citep{calders2010three}
\begin{align}
    \text{MD}(f) = P(\Yhat = 1 | A = 0) - P(\Yhat = 1 | A = 1).
\end{align}
and the disparate impact (DI) factor computes their ratio \citep{feldman2015certifying}.
%
Similar measures can be defined for equal opportunity and equalized odds.
For continuous variables, $\chi^2$-divergence measures the independence as stipulated by separation and independence \citep{mary2019fairness}.

Given a dataset $\Dcal_{tr}$ with $n$ examples, 
we can empirically estimate both sides of \eqref{eq:def_dem_par} by, 
\eg, for $a\in \{0,1\}$
\begin{align}
\label{eq:mean_est_DP}
    P(\Yhat = 1 | A = a)
    &\approx \text{mean} (\{ P_f(\Yhat = 1 | x_i) : a_i = a \}) \!\!\!\!\! \\
\nonumber
    &= \sum\nolimits_{i: a_i = a} P_f(\Yhat = 1 | x_i) \Big / \sum\nolimits_{i: a_i = a} \! 1.
\end{align}
Here, for a finite set $S$,
we denote the mean of its elements as $\text{mean} \, (S)$.
The above estimator is straightforward from Eq 1 and 7 of \citet{menon2018cost} in the setting of randomized classifier.
It is asymptotically unbiased, consistent, 
and \textbf{differentiable} in $f$.
see the proof in Appendix \ref{sec:property_estimator}.
Similarly, we can estimate
$
    P(\Yhat = 1 | A = a, Y = y) \approx \text{mean}\, (S), 
    \where S := \{ P_f(\Yhat = 1 | x_i) : a_i = a, y_i = y \}.
$
Other estimators can also be adopted, 
such as kernel density estimation \citep{Cho2020fair}.
Applying these estimates to MD, DI, $\chi^2$-divergence, or any other metric,
we obtain a fairness risk $\Fcal(P_f, \yvec, \avec)$.
We keep its expression general,
and some concrete examples are given in 
\eqref{eq:def_deo} to \eqref{eq:def_dpp} in Appendix~\ref{sec:multiclass_fairness}.
An \textbf{example} based on \eqref{eq:mean_est_DP} is
\begin{align*}
    \Fcal(P_f, \yvec, \avec)
    &:= | \text{mean} (\{ P_f(\Yhat = 1 | x_i) : a_i = 1 \}) \\
    &\qquad 
    - \text{mean} (\{ P_f(\Yhat = 1 | x_i) : a_i = 0 \}) |.
\end{align*}


\paragraph{Classification risk.}
We denote the conventional classification risk as $\Rcal(P_f, \yvec)$.
For example, the standard cross-entropy loss yields $- \frac{1}{n} \sum_i \log P_f(\Yhat = y_i | X = x_i)$.
%
%
%
%
Combined with fairness risk,
we can next find the optimal $f$
by  minimizing their sum weighted by $\lambda > 0$:
\begin{align}
\label{eq:fair_reg_classification}
    \Fcal(P_f, \yvec, \avec) + \lambda \cdot \Rcal(P_f, \yvec).
\end{align}
In \citet{williamson2019fairness}, a more general framework of fairness risk is constructed,
incorporating the loss $\ell$ into the definition of $\Fcal$ itself \citep{lahoti2020fairness}.
Our method developed below can be applied  directly.

\section{Fairness with  Group-Conditionally Unavailable Demographics}
\label{sec:risk_model}

The regularized objective \eqref{eq:fair_reg_classification} requires the demographic features which can often become unavailable due to a respondent's preference, privacy, and legal reasons.
As a key contribution of this work, 
we further address \textbf{group-conditionally} unavailable demographics,
where demographics can get unavailable in a \textit{non-uniform} fashion,
depending on the specific group.
This matches reality. 
For example, people above some age may be more reluctant to reveal it when applying entry-level jobs.

We show that this new setting can be approached by extending the objective \eqref{eq:fair_reg_classification}.
To this end,
we introduce a new random variable $\Atil$, 
which is the observation of the true latent demographic $A$.
For a principled treatment,
we treat $A$ as \textbf{never observed},
while $\Atil$ can be equal to $A$, 
or take a distorted value, 
or be unavailable (denoted as $\emptyset$).%
\footnote{We intentionally term it ``unavailable" instead of ``missing", 
because \textit{missing} means unobserved in statistics,
while we take \textit{unavailable} (i.e., $\emptyset$) as an \textit{observed} outcome.}
In other words, $\Atil$ is \textbf{always observed}, 
including taking the value of $\emptyset$.
As Section~\ref{sec:risk_model_vae} shows,
this novel model offers substantial flexibility,
even if the demographic contains multiple features and a different subset of them is unavailable for different individuals.

We model the group-conditional unavailability with a learnable noising probability $P(\Atil|A)$ \citep{Yao2023Latent}.
As a special case,
one may assert that demographics cannot be misrepresented, 
i.e., $P(\Atil|A) = 0$ for all $\Atil \notin \{A, \emptyset\}$.
%
In order to address isometry (e.g., totally swapping the concept of male and female), 
we impose some prior on $P(\Atil|A)$.
For example, $P(\Atil|A)$ must be low for $\Atil \notin \{A, \emptyset\}$.
This can be enforced by a Dirichlet prior,
e.g.,
$(P(\text{male} | \text{male}), P(\text{female} | \text{male}), P(\emptyset | \text{male})) \sim \text{Dir}(0.5, 0.1, 0.4)$.
%
Similarly, we can define $P(\Ytil|Y)$ for class-conditionally unavailable label,
allowing $\Ytil = \emptyset$.

In the sequel, we will assume there is an SSL algorithm that predicts $Y$ and $A$ with probabilistic models $P_f(Y|X, \Ytil)$ and $q_\phi(A|X, \Atil)$, respectively.
For example, the encoder of a VAE, 
which will be detailed in Section~\ref{sec:risk_model_vae} as \eqref{eq:q(a|xa)}.
%
%
Both $A$ and $Y$ can be \textbf{multi-class}.
It is now natural to extend the fairness risk by taking the \textit{expectation} of these missing values:
\begin{align}
\label{eq:def_frisk_expexp}
&\Ecal_{\text{vanilla}}(P_f, q_\phi) \\
\nonumber
    &:= \expunder{Y_i \sim P_f(Y|x_i, \ytil_i) }     
    \expunder{A_i \sim q_\phi(A | x_i, \atil_i) } \Fcal(P_f, Y_{1:n}, A_{1:n}).
\end{align}
%
In the case of no misrepresentation,
the expectation of $A_i$ can be replaced by $A_i = \atil_i$ if $\atil_i \neq \emptyset$.
Likewise for $Y_i$.
Adding this vanilla semi-supervised regularizer to an existing learning objective of $P_f$ and $q_\phi(A|X,\Atil)$ such as VAE gives a straightforward promotion of fairness,
and we will also discuss its optimization in Section~\ref{sec:gumbel_vanilla}.
However, despite its clear motivation,
we now demonstrate that it is indeed plagued with conceptual and practical issues, 
which we will address next.

\vspace{-1.2em}
\subsection{Rationalizing semi-supervised fairness risk}

It is important to note that the onus of fairness is only supposed to be on the classifier $P_f$,
while $q_\phi(A|X,\Atil)$---which infers the demographics to define the fairness metric itself---is supposed to be \textit{recused} from fairness risk minimization.
Reducing the risk by manipulating an individual's gender is not reasonable.
This issue does not exist in a fully supervised setting, 
but complicates the regularizer when $q_\phi(A|X,\Atil)$ is \emph{jointly} optimized with the classifier $P_f$.
For example, although the risk can be reduced by making $A$ independent of $X$,
this should not be intentionally pursued unless the underlying data distribution supports it.
The posterior $q_\phi(A|X,\Atil)$ is only supposed to accurately estimate the demographics,
while the task of fairness should be left to the classifier $P_f$.

A natural workaround is a two-step approach: 
first train a semi-supervised learner for the distribution $A|X, \Atil$,
and denote it as $r(A|X, \Atil)$.
Then in \eqref{eq:def_frisk_expexp},
we replace $A_i \sim q_\phi(A | x_i, \atil_i)$ with $A_i \sim r(A | x_i, \atil_i)$.
Although this resolves the above issue,
it decouples the learning of $A|X, \Atil$ (first step) and $Y|X, \Ytil$ (second step).
This is not ideal because, as commonly recognized, the inference of $A$ and $Y$ can benefit from \textit{shared} backbone feature extractors on $X$ (\eg, ResNet for images),
followed by finetuning target-specific heads.

However, sharing backbones will allow $q_\phi(A|X,\Atil)$ to be influenced by the learning of $P_f(Y|X, \Ytil)$,
conflicting with the aforementioned recusal principle.
In light of this difficulty, we resort to the commonly used stop-gradient technique that is available in PyTorch (\texttt{detach}) and TensorFlow (\texttt{tf.stop\_gradient}). 
While they still share backbones,
the derivative of $\Ecal_{\text{vanilla}}$ in \eqref{eq:def_frisk_expexp} with respect to $q_\phi(A|X,\Atil)$ is stopped (set to 0) from backpropagation,
\ie, treating $q_\phi(A|X,\Atil)$ as a constant.
It also much simplifies the training process. \highlight{The stop-gradient method is a technique that blocks the flow of gradient through a specific part of the computational graph during backpropagation, which is commonly used to decouple certain computations from parameter updates.}
\highlight{We present the corresponding results in our ablation study, shown by the blue curve among all experiment figures.}

\subsection{Imputation of unavailable training labels}

A similar issue is also present in the expectation $Y_i \sim P_f(Y|x_i, \ytil_i)$ in \eqref{eq:def_frisk_expexp}.
The ground truth label $Y_{1:n}$ is given in the supervised setting,
but its missing values should \textit{not} be inferred to account for fairness,
because labels are part of the fairness definition itself,
except for demographic parity.
We could resort to the stop-gradient technique again to withhold the backpropagation through $Y_i \sim P_f(Y|x_i, \ytil_i)$.
However, different from unavailable demographics, the situation here is more intricate because $P_f$ also serves as the first argument of $\Fcal(P_f, Y_{1:n}, A_{1:n})$,
where it indeed enforces fair classification.
This conflicts with the fairness-oblivious requirement for imputing $Y_i$,
which is also based on $P_f$.
In other words, we cannot enforce $P_f$ to both respect and disregard fairness at the same time.


To address this issue, we will train a \textit{separate} semi-supervised  
classifier to impute $Y_i$ without fairness concerns,
and it does leverage the backbone features mentioned above.
Suppose $\hvec_i$ is the current feature representation for $x_i$,
e.g., the third last layer of $f$.
Assuming there is no misrepresentation in $\ytil_i$,
we can clamp $Y|x_i$ to $P_f$ for those examples with observed label ($\ytil_i \neq \emptyset$).
Then we infer a class probability ${\color{red}P_g}(Y|x_i)$ for the rest examples by using $\hvec_i$ and any SSL algorithm such as graph-based Gaussian field \citep{ZhuLafGha03}.
%
This amounts to a fairness risk as

\vspace{-1.8em}
\begin{align}
\label{eq:def_frisk_final}
&\boxed{\Ecal(P_f, q_\phi)} \\
\nonumber
&:= \!\!
\expunder{\{Y_i \sim {\color{red} P_g}(Y|x_i): \ytil_i = \emptyset\}}     
    \expunder{A_i \sim {\color{red} q_\phi} (A | x_i, \atil_i)} 
    \!\! \Fcal(P_f, Y_{1:n}, A_{1:n}).
\end{align}
Note we stop the gradient on $P_g$ and $q_\phi$ here,
and $P_g$ is used instead of $P_f$.
Appendix~\ref{sec:ext_class_prob_app} extends this risk from randomized classifier to class-probability
estimator.


\begin{figure*}[t]
    \centering
    \begin{subfigure}[t]{0.48\textwidth}
        \centering
        \includegraphics[width=0.6\linewidth]{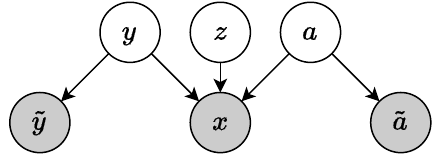}
        \vspace{-0.45em}
        \caption{Decoder}
        \label{fig:decoder}
    \end{subfigure}
    ~~~~
    \begin{subfigure}[t]{0.48\textwidth}
        \centering
        \includegraphics[width=0.6\linewidth]{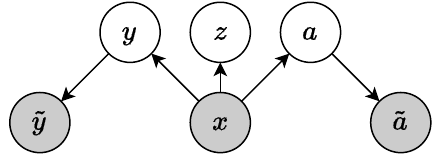}
        \vspace{-0.45em}
        \caption{Encoder}
        \label{fig:encoder}
    \end{subfigure}
    \vspace{-0.5em}
    \caption{SS-VAE decoder and encoder with unavailable demographic/label conditioned on group/class}
    \label{fig:conditional_vae}
    \vspace{-0.5em}
\end{figure*}

\subsection{Efficient evaluation of fairness risk $\Ecal$}
\label{sec:monte_carlo}

A major challenge in risk-based methods is the cost of computing the expectations in $\Ecal$,
along with its derivatives.
In the sequel, we will resort to a Monte Carlo based method,
such that for $n$ training examples,
an $\epsilon$ accurate approximation with confidence $1-\delta$ can be found with  $O(\frac{c_n}{\epsilon^2} \log \frac{1}{\delta})$ computation,
where $c_n$ is the cost of evaluating $\Fcal$ and it is $O(n)$ in our considered cases.
Although some customized algorithms can be developed by exploiting the specific structures in $\Fcal$ such as demographic parity,
we prefer a more general approach.
Despite the inevitable inexactness in the result, 
we prove tight concentration bounds that turn out sufficiently accurate in practice.

Our method simply draws $N$ number of iid samples from $A_i \sim q_\phi(A|x_i, \atil_i)$ and $Y_i \sim P_g(Y|x_i)$.
For $s = 1, \ldots, N$,
let $Z_s := \{a^{(s)}_i, y^{(s)}_i\}_{i=1}^n$.
We estimate $\Ecal$ by
\begin{align}
\label{eq:def_MC}
    \hat{\Ecal}_n(Z_1, \ldots, Z_N) := \smallfrac{1}{N}
    \sum\nolimits_{s=1}^N \Fcal(P_f, Z_s).
\end{align}
We prove the following sample complexity for this estimator by leveraging McDiarmid's inequality.

\begin{theorem}
\label{thm:bound_sample}
    Suppose $\Fcal \in [0, C]$ where $C > 0$ is a constant.
    Then for all $\epsilon > 0$,
    \begin{align*}
        P(|\hat{\Ecal}_n(Z_1, \ldots, Z_N) - \Ecal| \ge \epsilon) 
    \le \ 2 \exp(- 2 N \epsilon^2 / C^2).
    \end{align*}
As a result, to guarantee an estimation error of $\epsilon$ with confidence $1-\delta$,
it suffices to draw $N = \frac{C^2}{2 \epsilon^2} \log \frac{1}{\delta}$ samples.
The proof is available in Appendix \ref{sec:proof_sample_app}.
\end{theorem}

\highlight{This theorem is non-trivial because the fairness risk can depend on the predicted probabilities in a complex manner, creating nonlinear dependencies among all the training examples; see Appendix \ref{sec:multiclass_fairness}. The analysis also reveals the limitations of the sampler, as certain risks, such as disparate impact, are unbounded.}

As $\hat{\Ecal}_n(Z_1, \ldots, Z_N)$ costs $O(nN)$ to compute,
the total cost is $\frac{n C^2}{2 \epsilon^2} \log \frac{2}{\delta}$.
When $C=1$, $\epsilon = 0.01$ and $n = 10^4$, 
this is order of $10^8$,
which is quite affordable.
In experiments, a sample size \underline{\textbf{as low as}} \underline{\bf{$N = 100$}} was sufficient for our method to outperform the state of the art; 
see Section~\ref{sec:experiment}.
Most risks we consider satisfy the boundedness assumption,
although exceptions exist such as disparate impact.

\vspace{-0.5em}
\subsection{Differentiation of vanilla fairness risk}
\label{sec:gumbel_vanilla}

Since we stop the gradient with respect to $q_\phi(A | x_i, \atil_i)$ and $P_g(Y|x_i)$,
the whole regularizer $\Ecal(P_f, q_\phi)$ in \eqref{eq:def_frisk_final} can be easily differentiated with respect to $f$.

In practice, we also would like to conduct an ablation study by comparing with $\Ecal_{\text{vanilla}}$,
where challenges arise from differentiation with respect to $q_\phi(A | x_i, \atil_i)$ and $y_i \sim P_f(Y|x_i, \ytil_i)$.
%
We resolve this issue by using the straight-through Gumbel-Softmax method 
\citep{Jang2017categorical,maddison2017the}.
A self-contained description is given in Appendix~\ref{sec:gumble_app}.

\vspace{-0.5em}
\section{Integrating Fairness Risk with SSL}
\label{sec:risk_model_vae}
\vspace{-0.1em}




We next illustrate how the fairness risk can be integrated with SSL.
As an example, we recap the semi-supervised VAE \citep[SS-VAE,][]{Kingma2014semi} with missing $a$ and $y$ values.%
\footnote{It is customary in statistics to denote random variables by capital letters.
However, the VAE literature generally uses lowercase letters.
We thus switch to their custom.
}

\highlight{\paragraph{Why VAE?}
In terms of SSL, while some discriminative methods exist such as \citet{ZhuLafGha03}, the most prevalent and effective approaches are based on generative models, such as SS-VAE. 
Indeed, most conventional generative models for SSL are essentially VAEs, especially when they are trained with variational inference.

Secondly, our approach to fairness risk relies on an encoder capable of estimating the posterior probability of missing labels or demographics.
This rules out normalizing flows or generative adversarial networks unsuitable, as the latter’s discriminator does not serve as an encoder. 
Moreover, transformers are not considered since our formulation does not involve sequential data modeling.

In contrast, VAE employs an encoder and provides the conditional independencies that allow us to more finely specify the constituent distributions as in \eqref{eq:decoder_def} and \eqref{eq:q(a|xa)}. Diffusion models, as a more refined variant of VAE, can also be adopted. However, we intentionally keep the generative model simple, prioritizing fairness modeling, while future work can incorporate additional refinements to the VAE.}
%

\vspace{-0.5em}
\subsection{Encoders and decoders}

SS-VAE employs a decoder/generation process and an encoder/inference process parameterized by $\theta$ and $\phi$ respectively. 
In the decoder whose graphical model is shown in Figure \ref{fig:decoder},
the observation $\xvec$ is generated conditioned on the latent variable $\zvec$, 
latent class $y$ and latent demographics $a$:

\vspace{-2.05em}
\begin{align}
\nonumber
    p_\theta(\xvec, \atil, \ytil|a, y, \zvec) &= \Ncal(\xvec| g_\theta(a, y, \zvec)) \cdot P_\theta(\atil|a) \cdot P_\theta(\ytil|y), \\
\nonumber        
    p(y) &= \text{Cat}(y|\pi_y),\quad
    p(\pi_y) = \text{SymDir}(\gamma_y), \\
\nonumber
    p(a) &= \text{Cat}(a|\pi_a), \quad
    p(\pi_a) = \text{SymDir}(\gamma_a), \\
\label{eq:decoder_def}
    p(\zvec) &= \Ncal(\zvec|0, I).
\end{align}
Here $\text{Cat}(y|\pi_y)$ is the multinoulli  prior of the class variable.
SymDir is the symmetric Dirichlet with hyperparameter $\gamma_y$.
The prior of $a$ is analogously defined.

The inference process finds conditionally independent representations
$\zvec$, $a$, and $y$ under a given $\xvec$, $\atil$, and $\ytil$. 
Accordingly, the approximate posterior/encoder $q_\phi(a, y, \zvec| \xvec, \atil, \ytil)$ factors as in Figure~\ref{fig:encoder}:

\vspace{-2.05em}
\begin{align}
\nonumber
    q_\phi(a, y, \zvec| \xvec, \atil, \ytil) &= q_\phi(\zvec|\xvec) \cdot q_\phi(y|\xvec, \ytil)  \cdot q_\phi(a|\xvec, \atil) 
    \qquad \\
\nonumber    
    \text{where} \ \  
    \quad q_\phi(\zvec|\xvec) &= \Ncal(\zvec|\mu_\phi(\xvec), \sigma^2_\phi(\xvec)), \quad \\
\nonumber    
    q_\phi(y|\xvec, \ytil) &\propto q_\phi(y|\xvec) \cdot P_\theta(\ytil|y) \\
\nonumber
    &= \text{Cat}(y|g_\phi(\xvec)) \cdot P_\theta(\ytil|y), \\
\nonumber
\label{eq:q(a|xa)}
    q_\phi(a|\xvec, \atil) &\propto q_\phi(a|\xvec) \cdot P_\theta(\atil|a) \\
    &= \text{Cat}(a|h_\phi(\xvec)) \cdot P_\theta(\atil|a).
\end{align}
Here $\mu_\phi, \sigma_\phi, g_\phi, h_\phi$ are all defined as neural networks. 
Compared with Figure~\ref{fig:decoder},
here we only reversed the arrows connected with $\xvec$,
following a standard assumption in VAE that, in posterior, $y$, $z$, $a$ are independent given $\xvec$.
We maximally preserved the other arrows, 
namely $y \to \ytil$ and $a \to \atil$.
The chain of $\xvec \to a \to \atil$ can be interpreted as first probabilistically determining $a$ based on $\xvec$,
and then adding noise to it producing $\atil$ (including $\emptyset$). 
This chain structure allows us to derive $q_\phi(a|\xvec, \atil)$ in \eqref{eq:q(a|xa)} (and analogously $q_\phi(y|\xvec, \ytil)$) as:
\begin{align*}
    p(a|\xvec, \atil) &\propto p(a) p(\xvec, \atil|a) = p(a) p(\xvec|a) p(\atil|a) \\   
    &= p(\xvec) p(a|\xvec) p(\atil|a) 
     \propto p(a|\xvec) p(\atil|a).
\end{align*}
%
%
\textbf{In training}, the fairness risks in \eqref{eq:def_frisk_expexp} and \eqref{eq:def_frisk_final} require $P_f(y|\xvec,\ytil)$,
which can be  served by $q_\phi(y|\xvec, \ytil)$.

\subsection{Instilling Fairness to SS-VAE}
\label{sec:fair_VAE}

The evidence lower bound (ELBO) of $\log p_\theta(\xvec, \atil, \ytil)$ can be derived in a standard fashion,
and we relegate the details to Appendix 
\ref{sec:app_ELBO}.
Denoting it as $\text{ELBO}(\xvec,\atil,\ytil)$,
we extend the SS-VAE objective as follows,
to be minimized over $\theta$ and $\phi$:
\begin{align}
\nonumber
    &\Lcal(\theta, \phi) = 
    \Omega(P_\theta(\atil|a), P_\theta(\ytil|y)) 
    \\
\label{eq:obj_VAE}
&\qquad \qquad - \expunder{(\xvec,\atil, \ytil) \sim  \Dcal_{tr}} 
    \!\! \Big[ \text{ELBO}(\xvec,\atil,\ytil)\\
\nonumber
    &\qquad \qquad \qquad + \sembrack{\atil \neq \emptyset} \cdot \log \sum\nolimits_{a} \text{Cat}(a|h_\phi(\xvec)) P_\theta(\atil|a) \\
\nonumber
    &\qquad \qquad \qquad +\sembrack{\ytil \neq \emptyset} \cdot \log \sum\nolimits_{y} \!\! 
    \text{Cat}(y|g_\phi(\xvec)) P_\theta(\ytil|y) \Big].
\end{align}
Here, the regularizer $\Omega$ enforces the Dirichlet prior. 
For example, let 
$a = P(\text{male}|\text{male})$, 
$b = P(\text{female}|\text{male})$, 
$c = P(\emptyset|\text{male})$. 
Then a prior of $\text{Dir}(0.5, 0.2, 0.4)$ leads to a regularizer of $\log (a^{0.5-1} b^{0.2-1} c^{0.4-1})$. 
\highlight{The regularization is beneficial when prior knowledge about the distributions is available, such as the likelihood that a male is more likely to be labeled as male rather than female. However our experiments did not make any such assumptions and thus did not incorporate this regularizer.}
Suppose $\ytil \neq \emptyset$.  
If the provided $\ytil$ always truthfully represents $y$,
then $P(\ytil | y) = \sembrack{\ytil = y}$.
As a result, 
the last line in \eqref{eq:obj_VAE} equals $\text{Cat}(\ytil|g_\phi(\xvec))$,
the standard supervised loss in SS-VAE.

Casting SS-VAE into the fairness risk framework,
the role of $f$ is played by $q_\phi(y|\xvec)$ and we only need to augment the objective \eqref{eq:obj_VAE} into
\begin{align}
\label{train_objective}
\boxed{
\Lcal(\theta, \phi) + \lambda \cdot \Ecal(q_\phi(y|\xvec), q_\phi(a|\xvec)),}
\end{align}
where $\lambda >0$ is a tradeoff hyperparameter \highlight{and $\Ecal$ takes the same form as in \eqref{eq:def_frisk_final}}.
We will henceforth refer to this model as \textbf{Fair-SS-VAE}.

\paragraph{Classifying test data.}

We simply predict by 
\begin{align}
\label{test_objective}
P_f(y|\xvec) := q_\phi(y|\xvec) = \text{Cat}(y|g_\phi(\xvec)),
\end{align}
with no access to $\atil$ \citep{lipton2018does}.
%
A more principled approach is to minimize the risk of fairness and classification,
which is relegated to Appendix~\ref{sec:inf_risk_app}.

Note our new VAE can readily extend $y$ and $a$ from binary values to \textbf{multi-class} and \textbf{continuous},
with little impact on Monte Carlo sampling in \eqref{eq:def_MC}.
So our Fair-SS-VAE is applicable wherever the underlying fairness risk is available;
see Appendix~\ref{sec:multiclass_fairness}.

\begin{figure*}[t]
\begin{subfigure}{0.323\textwidth}
  \centering
  \includegraphics[width=\linewidth,clip=true, viewport = 0.4cm 0.1cm 16.3cm 14.1cm]{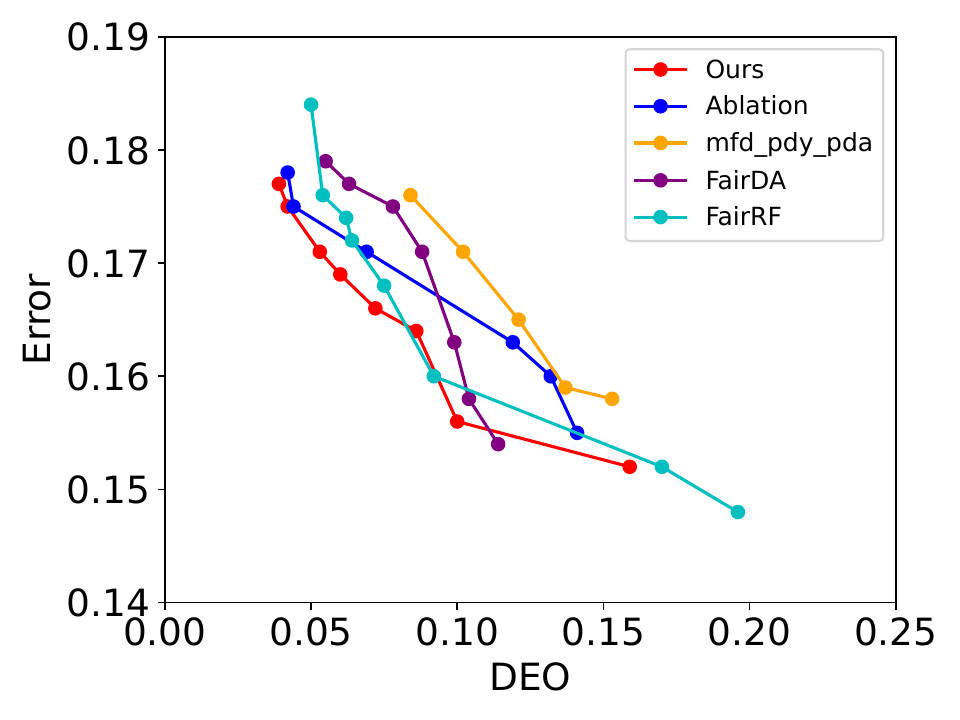}\quad
  \caption{DEO-Sparse}
\end{subfigure}
\begin{subfigure}{0.323\textwidth}
  \centering
  \includegraphics[width=\linewidth,clip=true, viewport = 0.4cm 0.1cm 16.3cm 14.1cm]{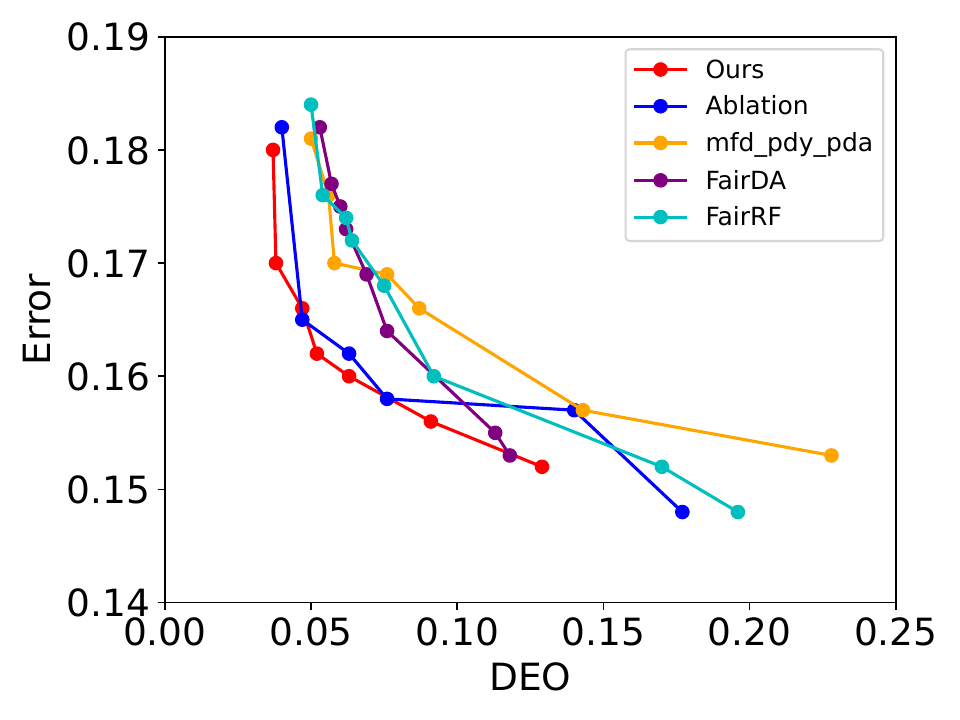}\quad
  \caption{DEO-Medium}
\end{subfigure}
\begin{subfigure}{0.323\textwidth}\quad
  \centering
  \includegraphics[width=\linewidth,clip=true, viewport = 0.4cm 0.1cm 16.3cm 14.1cm]{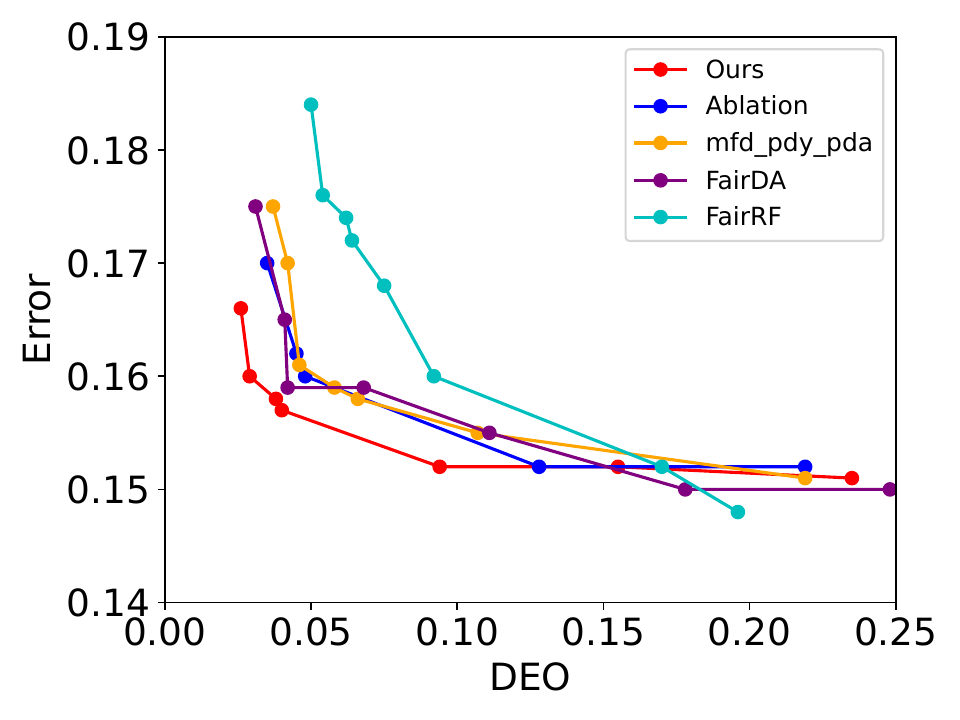}
  \caption{DEO-Dense}
\end{subfigure}
\\
  \centering
\begin{subfigure}{0.323\textwidth}
  \centering
  \includegraphics[width=\linewidth,clip=true, viewport = 0.3cm 0.1cm 17.1cm 14.1cm]{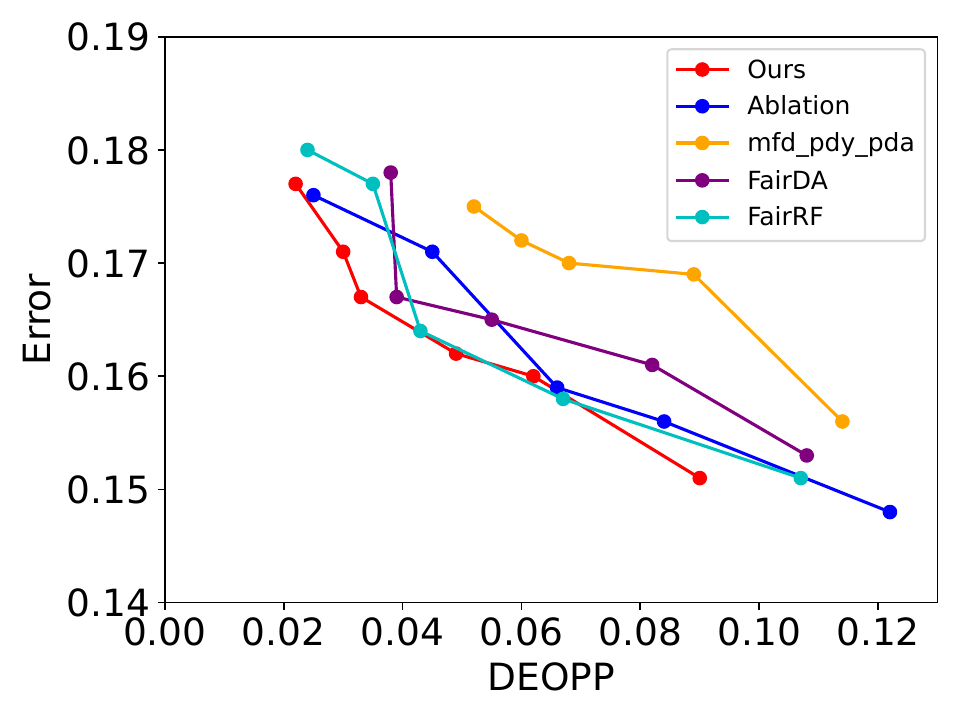}\quad
  \caption{DEOPP-Sparse}
\end{subfigure}
\begin{subfigure}{0.323\textwidth}
  \centering
  \includegraphics[width=\linewidth,clip=true, viewport = 0.3cm 0.1cm 17.1cm 14.1cm]{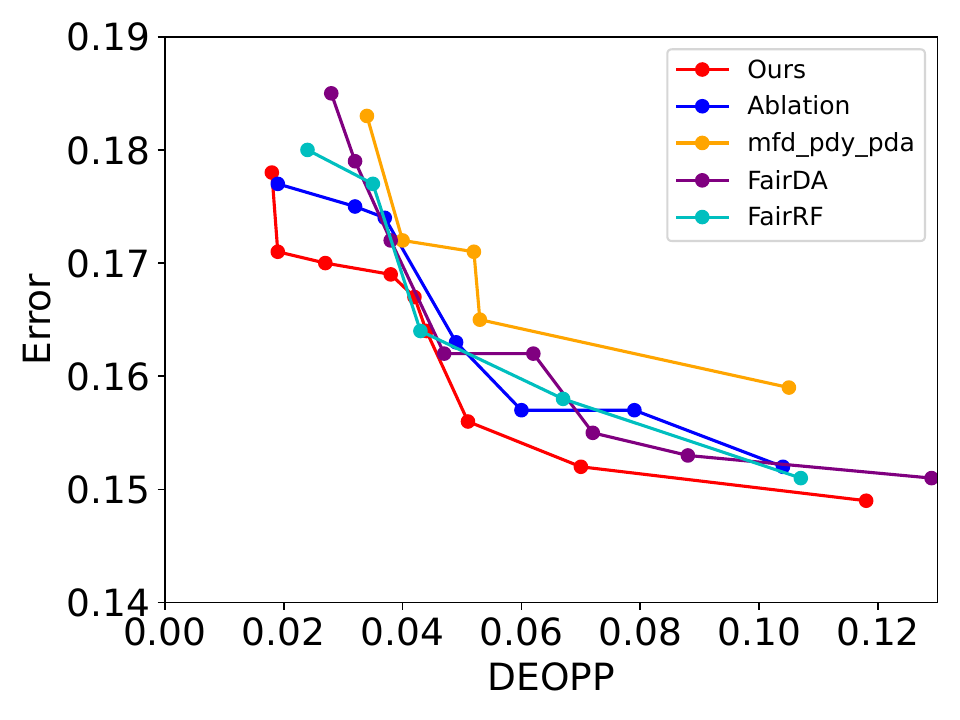}\quad
  \caption{DEOPP-Medium}
\end{subfigure}
\begin{subfigure}{0.323\textwidth}
  \centering
  \includegraphics[width=\linewidth,clip=true, viewport = 0.3cm 0.1cm 17.1cm 14.1cm]{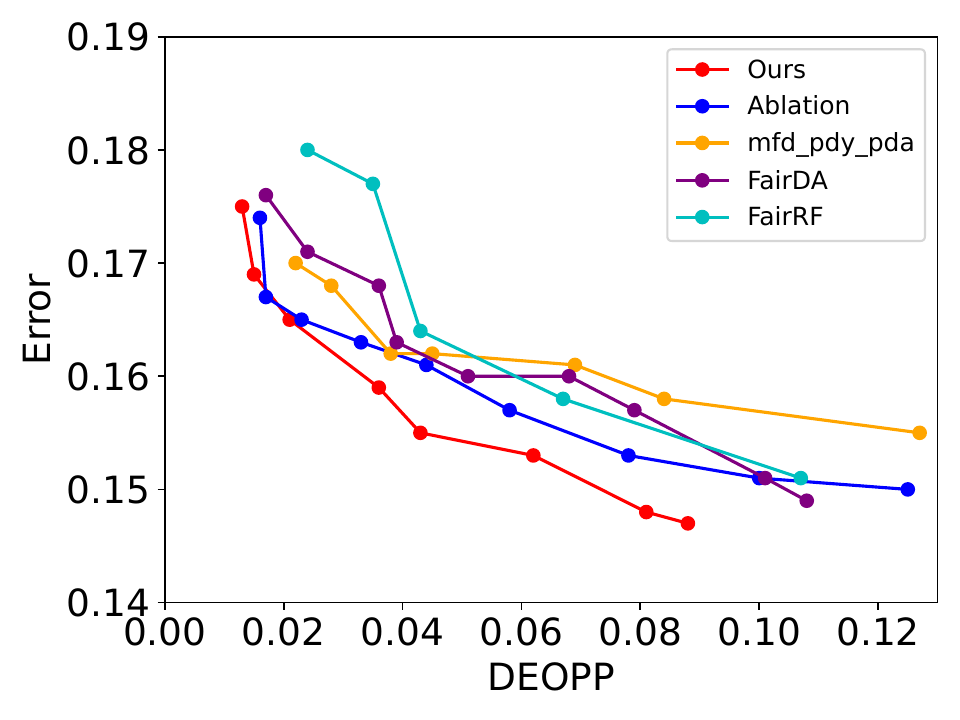}
  \caption{DEOPP-Dense}
\end{subfigure}
\vspace{-0.2em}
\caption{Pareto frontier of error versus DEO/DEOPP for \textbf{Adult-Gender}}
\label{fig:adult_gender}
\vspace{-1em}
\end{figure*}

\section{Experimental Results}
\label{sec:experiment}

We now show empirically that with group-conditionally missing demographics and generally missing labels, 
our Fair-SS-VAE significantly outperforms various state-of-the-art semi-supervised fair classification methods in terms of various fairness metrics,
while keeping a similar accuracy.

\vspace{-0.5em}
\paragraph{Datasets}
We used two datasets to create \textit{three} sets of experiments:

\textbf{CelebA} \citep{liu2015deep}. 
CelebA contains face images annotated with 40 binary attributes. 
We sampled 45k images as our training and validation dataset, and 5k as test set. 
Following \citep{jung2022learning},
we set "Attractive" as the target label, and "Gender" as the sensitive group attribute. 


\textbf{UCI Adult} \citep{misc_adult_2}. 
The Adult dataset is a tabular dataset where the target label is whether the income exceeds \$50K per year given a person's attributes. 
We conducted two experiments with \textbf{gender} and \textbf{race} serving the sensitive attribute. 
The same pre-processing routine as in \citet{bellamy2018ai} was adopted.

\highlight{It is worth mentioning that conducting experiments where there is a significant correlation between target labels and groups is necessary. If the labels and groups are already uncorrelated, fairness modeling is likely less critical. 
For example, in CelebA, when considering a label/group pair with very low correlation, such as 'Eyeglasses' and 'Male', the model already achieves high accuracy with a DEO close to zero (ranging from 0.000 to 0.003) even without any fairness interventions. This suggests that fairness-aware models, including ours, have limited potential for improvement, making them less ideal for experiments.}

To simulate the missing data,
we randomly masked 25\% labels,
and masked the demographics as:
\begin{align*}
\nonumber
    P(\Atil \! = \! \emptyset | A \! =\! 1) = \alpha, \ \ 
    P(\Atil \! = \! 1 | A \! = \! 1) = 1 \! - \! \alpha, \\
    P(\Atil \! = \! \emptyset | A \! = \! 0) = \beta, \ \ 
    P(\Atil \! = \! 0 | A \! = \! 0) = 1 \! - \! \beta.
\end{align*}
We considered three levels of missing demographics: 
\textbf{sparse} ($\alpha = 0.4, \beta = 0.8$), 
\textbf{medium} ($\alpha = 0.2, \beta = 0.4$), and 
\textbf{dense} ($\alpha = 0.1, \beta = 0.2$).
Fair-SS-VAE can infer $\alpha$ and $\beta$ from the data.
Each dataset was first randomly partitioned into training, validation, and testing. 
Masking was applied to the first two,
and no group information is used in testing.

\begin{figure*}[t]
\begin{subfigure}{0.323\textwidth}
  \centering
  \includegraphics[width=\linewidth,clip=true, viewport = 0.4cm 0.1cm 16.4cm 14.1cm]{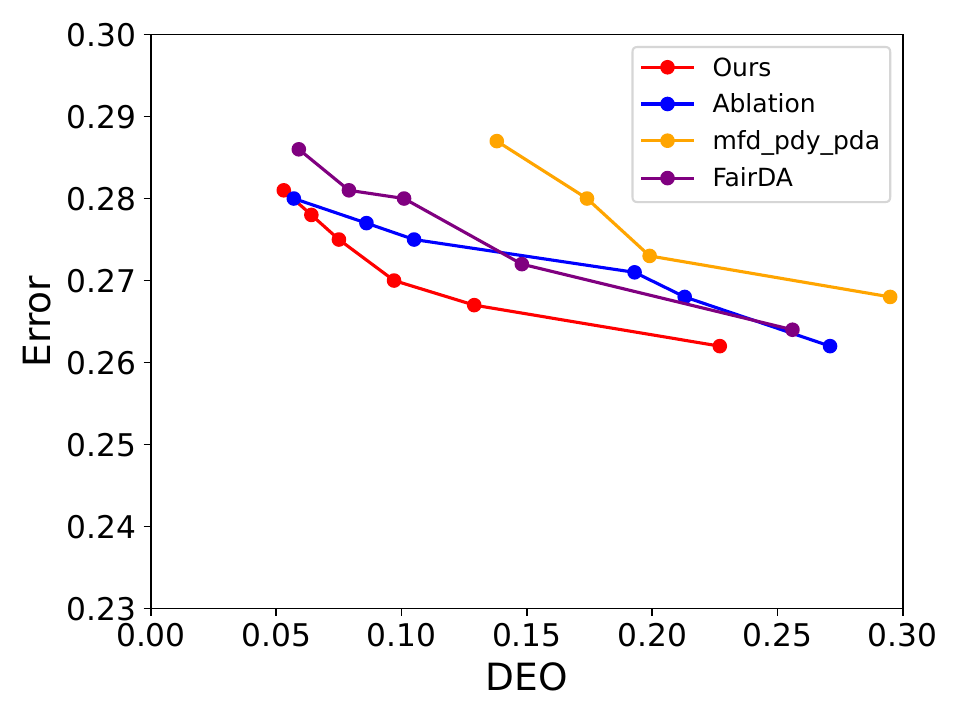}\quad
  \caption{DEO-Sparse}
\end{subfigure}
\begin{subfigure}{0.323\textwidth}
  \centering
  \includegraphics[width=\linewidth,clip=true, viewport = 0.4cm 0.1cm 16.4cm 14.1cm]{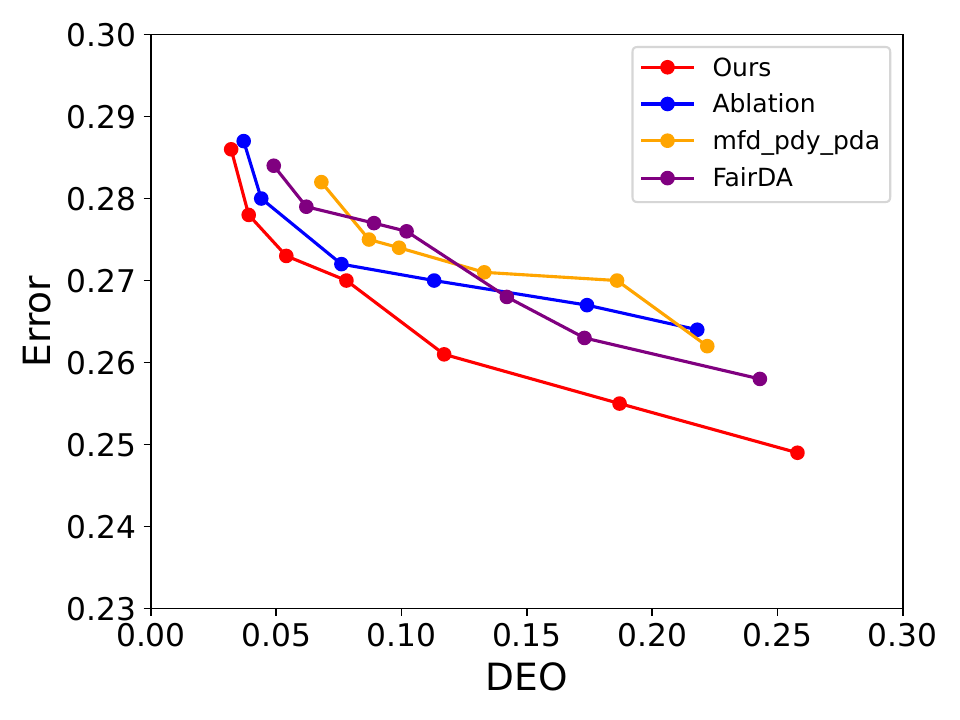}\quad
  \caption{DEO-Medium}
\end{subfigure}
\begin{subfigure}{0.323\textwidth}\quad
  \centering
  \includegraphics[width=\linewidth,clip=true, viewport = 0.4cm 0.1cm 16.4cm 14.1cm]{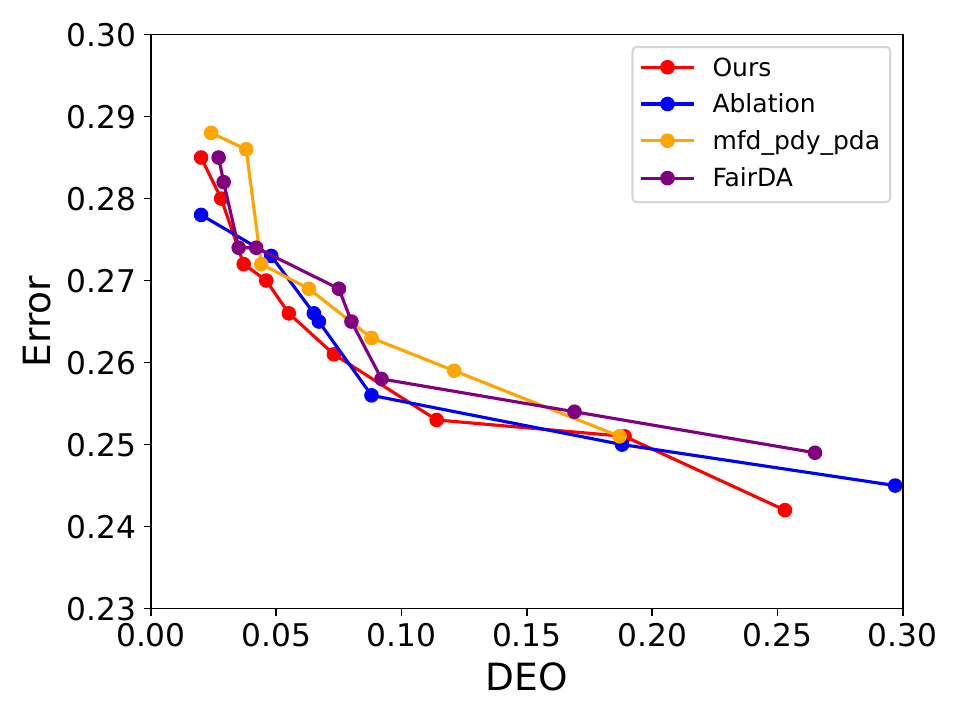}
  \caption{DEO-Dense}
\end{subfigure}
\\
  \centering
\begin{subfigure}{0.323\textwidth}
  \centering
  \includegraphics[width=\linewidth,clip=true, viewport = 0.3cm 0.1cm 17.1cm 14.1cm]{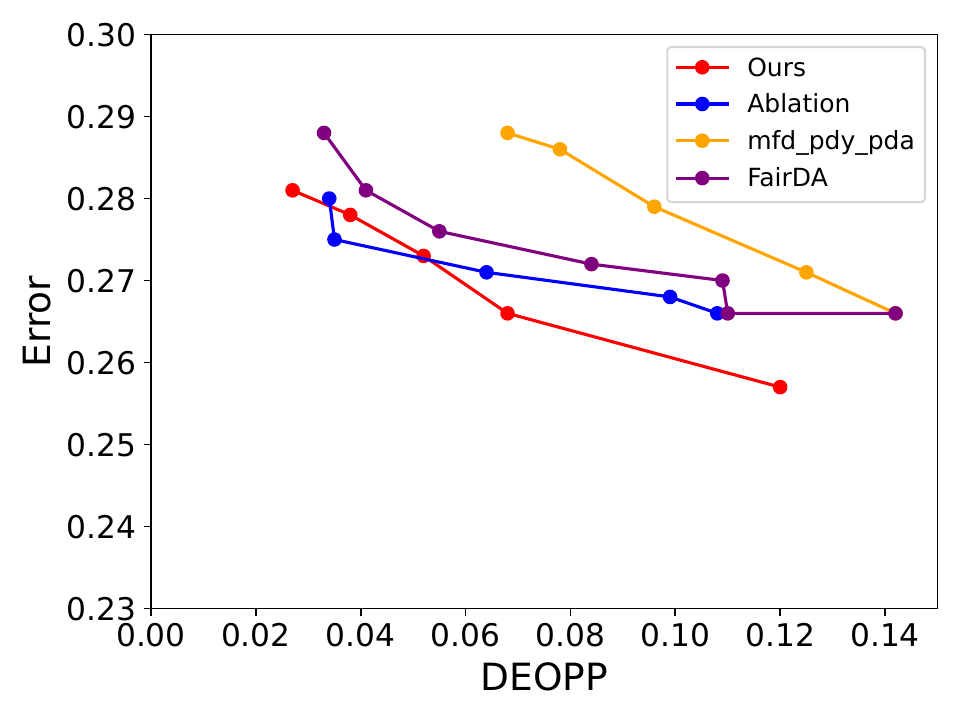}\quad
  \caption{DEOPP-Sparse}
\end{subfigure}
\begin{subfigure}{0.323\textwidth}
  \centering
  \includegraphics[width=\linewidth,clip=true, viewport = 0.3cm 0.1cm 17.1cm 14.1cm]{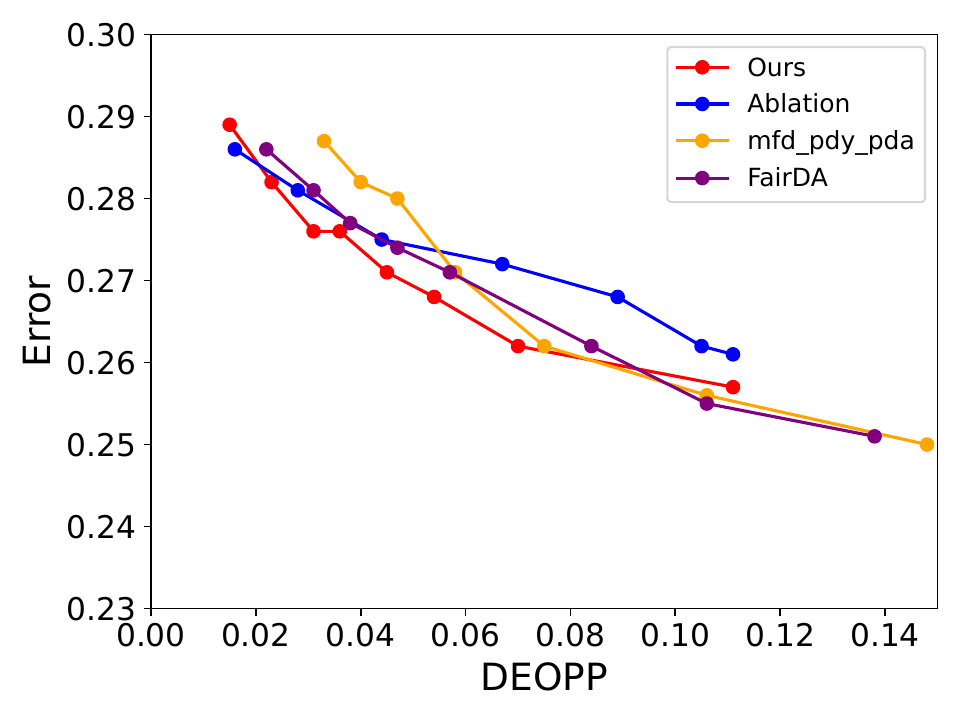}\quad
  \caption{DEOPP-Medium}
\end{subfigure}
\begin{subfigure}{0.323\textwidth}
  \centering
  \includegraphics[width=\linewidth,clip=true, viewport = 0.3cm 0.1cm 17.1cm 14.1cm]{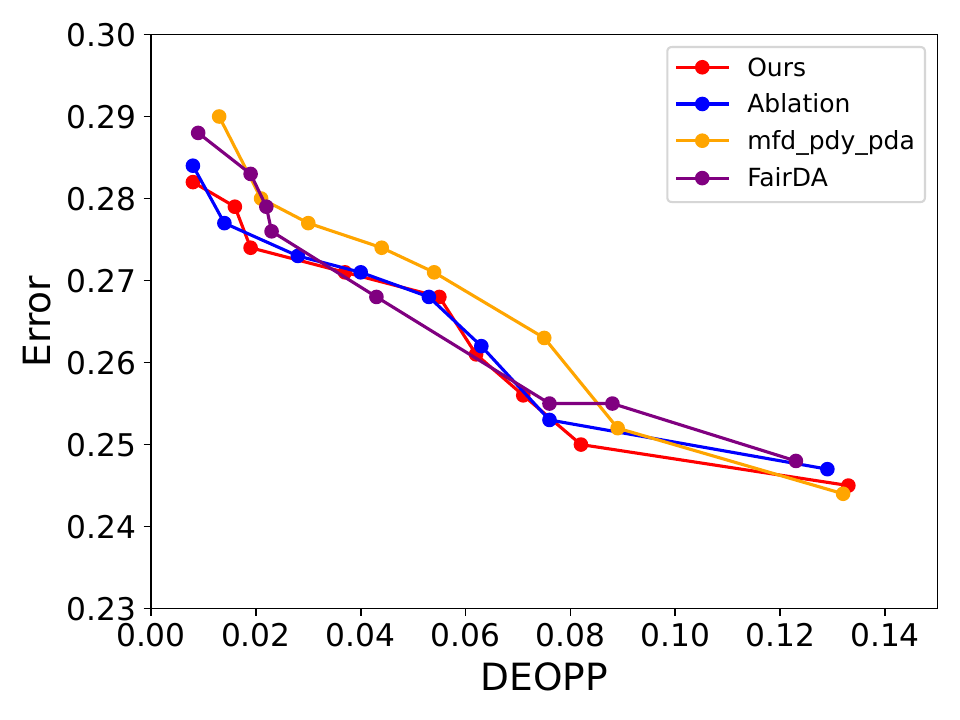}
  \caption{DEOPP-Dense}
\end{subfigure}
\vspace{-0.3em}
\caption{Pareto frontier of error versus DEO/DEOPP for \textbf{CelebA}}
\label{fig:celeba}
\vspace{-0.3em}
\end{figure*}

\paragraph{Baseline fair classifiers}

We compared with three types of methods.
%
MMD-based Fair Distillation \citep[\textbf{MFD},][]{jung2021fair}
encourages fairness through feature distillation,
but it requires observed labels and demographics.
So we imputed $y$ and $a$ with three heuristics: 
\begin{itemize}[leftmargin=*]
	\setlength{\itemsep}{2pt}
	\setlength{\parskip}{0pt}
    \vspace{-1em}
    \item (\textbf{gt y, gt a}): feed the ground-truth $y$ and $a$. It forms an unfair comparison with our method but sheds light on the best possible performance of the method;
    \item (\textbf{gt y, rand a}): feed the ground-truth $y$ and impute $a$ uniformly at random to eliminate the correlation between labels and groups;
    \item (\textbf{pred y, pred a}): train a classifier to predict the missing $y$ and $a$;
    \vspace{-1em}
\end{itemize}

As plotting three curves makes the figures overly crowded,
we defer the full plots to Appendix~\ref{sec:app_experiment},
and only show in the main paper the results without involving ground-truth imputation (the last variant).

The second baseline is \textbf{FairRF} \citep{ZhaDaiShuWan22},
which uses proxy features.
For Adult, they used \textit{age}, \textit{relation} and \textit{marital status} as the proxy features.  We found empirically that using \textit{age} alone gave even better results.  
So we just used \textit{age} as the proxy feature.
We did not apply FairRF to CelebA because proxy features are difficult to identify for images.

The third baseline is \textbf{FairDA} \citep{liang2023fair}.
It divides all examples into source and target domains based on the availability of sensitive feature,
and estimates that for the target domain by domain adaptation.


%


\vspace{-0.3em}
\paragraph{Implementation details of Fair-SS-VAE}

We employed M2-VAE \citep{Kingma2014semi} for Fair-SS-VAE,
with WideResNet-28-2 used as the feature extractor for CelebA,
and three fully connected layers for Adult.
We kept the sample size to 100 in the Monte-Carlo evaluation of $\Ecal$ in \eqref{eq:def_MC}.

\vspace{-0.3em}
\paragraph{Fairness performance metrics}

We measured the fairness of the predictions by \textit{difference of equalized odds} (\textbf{DEO}) and
\textit{difference of equal opportunity} (\textbf{DEOPP}).
On a test set, the prediction $P(\Yhat = 1 | x_i)$ is valued 0 or 1,
simply rounded from $q_\phi(y|x)$.
Surprisingly, it provides a very strong initialization for minimizing the expected fairness risk in \eqref{test_objective}.
In fact, it directly hits a local optimal,
leaving no improvement possible from coordinate descent.
This suggests that the fairness regularization in Fair-SS-VAE allows VAE to learn a posterior model that already accounts for the desired fairness,
dispensing with the need of risk minimization at test time.


\vspace{-0.3em}
\paragraph{Model selection} 

Since we will plot the Pareto frontier of classification error and fairness metrics,
ideally we only need a training and test set,
and to enumerate all the hyperparameter values,
based on which the Pareto frontier can be plotted.
However, doing so would be too expensive in computation.
Therefore, we used 10\% of the data to tune all the other hyper-parameters (network architecture, optimization, etc),
and varied the tradeoff weight $\lambda$ in \eqref{train_objective} to roll out the frontier.

\highlight{
Precisely, we focused on varying the tradeoff weight between accuracy and fairness while keeping other hyperparameters fixed, such as batch size, learning rate, and the weights for KL divergence and reconstruction error in the VAE.
Since these hyperparameters are already part of the standard SS-VAE, we relied on SS-VAE to select their values. After determining these values, we put back the fairness risk and further adjusted the selected hyperparameters, generating a new set of candidate values for model selection based on the validation set.
}


\subsection{Main results}

Figures~\ref{fig:adult_gender} and \ref{fig:celeba} show respectively the Pareto frontier of classification error and fairness metrics for Adult-Gender (with gender being the sensitive feature) and CelebA.
Due to space limitation, 
we defer the result of Adult-Race to Figure~\ref{fig:adult_race} in Appendix \ref{sec:app_experiment}.
As the missing demographics decrease from left to right (sparse to dense),
the Pareto Frontier tends to shift from the upper right corner to the bottom left corner, 
suggesting that all methods show improvement in the fairness metric (both DEO and DEOPP) while maintaining the same level of accuracy.

Fair-SS-VAE excels in producing a Pareto Frontier significantly closer to the origin across diverse datasets and varying levels of sparsity, 
indicating its superiority in balancing the tradeoff between accuracy and the fairness metric, unless the ground-truth $y$ and $a$ are used, which would make the comparison unfair to Fair-SS-VAE (see Appendix~\ref{sec:app_experiment}).
Overall, FairDA is the second most effective.

\vspace{-0.5em}
\paragraph{Recovery of group-conditional missing probability}

Table~\ref{tab:recovered_rate} presents the value of $\alpha$ and $\beta$ that Fair-SS-VAE finds for the three datasets and three sparsity levels.
Such a recovery is difficult because, once masked, the data does not carry the ground-truth of group feature, 
precluding counting for rate estimation.
Although the exact values are hard to recover, 
in many cases, they turn out close,
and the overall trend correctly decreases from sparse to dense.

\begin{table}[t!]
\setlength\tabcolsep{1.3pt}
\vspace{-0.7em}
    \caption{\centering Recovery rate of sensitive attribute by Fair-SS-VAE. 
    \textbf{Ad-G}: Adult-Gender, 
    \textbf{Ad-R}: Adult-Race, 
    \textbf{CelA}: CelebA}
    \label{tab:recovered_rate}
    \centering
    \begin{tabular}
    {c|cc|cc|cc}
        \toprule
            & \multicolumn{2}{c|}{Sparse}         
               & \multicolumn{2}{c|}{Medium}
               & \multicolumn{2}{c}{Dense}
\\    
         &  
        $\alpha = .4$ & 
        $\beta = .8$ &
        $\alpha = .2$ & 
        $\beta = .4$ &
        $\alpha = .1$ & 
        $\beta = .2$ \\
        \midrule
         Ad-G & .43{\footnotesize$\pm$.16} & .86{\footnotesize$\pm$.17} & .32{\footnotesize$\pm$.06} & .35{\footnotesize$\pm$.06} & .11{\footnotesize$\pm$.07} & .28{\footnotesize$\pm$.04}
          \\
          \midrule
         Ad-R & .39{\footnotesize$\pm$.23} & .82{\footnotesize$\pm$.15} & .30{\footnotesize$\pm$.04} & .42{\footnotesize$\pm$.13} & .17{\footnotesize$\pm$.01} & .25{\footnotesize$\pm$.12}
          \\
          \midrule
          CelA & .34{\footnotesize$\pm$.16} & .67{\footnotesize$\pm$.02} & .08{\footnotesize$\pm$.02} & .44{\footnotesize$\pm$.13} & .16{\footnotesize$\pm$.04} & .29{\footnotesize$\pm$.04}
          \\          
         \bottomrule
    \end{tabular}
     \vspace{-1em}      
\end{table}

\paragraph{Ablation study of stopping gradient}

To investigate the significance of stop gradient,
we set up an ablation experiment that directly uses $\Ecal_{\text{vanilla}}$. 
The corresponding results are labeled ``Ablation'' in Figure~\ref{fig:adult_gender}, \ref{fig:celeba} and \ref{fig:adult_race}.
Under most circumstances, it performs worse than Fair-SS-VAE, with occasional advantages when applied to datasets with dense sparsity levels. \highlight{Additionally, we carried out experiments titled CelebA-HighCheekbones, as illustrated in \ref{fig:celeba_highcheekbones}, where high cheekbones is the target label and gender serves as the group attribute.
}

\vspace{-0.4em}
\paragraph{Conclusion, limitation, broader impact, and future work}

We proposed a new fair classifier that addresses group-conditionally missing demographics.
Promising empirical performance is shown.
As a limitation,
we did not incorporate proxy features when they are available.
For future work, we will enable it by conditioning the VAEs on the proxy features.
We will also model the bias between \textit{different} sensitive groups, 
\eg, people of some \textit{race} are less reluctant to reveal their \textit{gender}.

\paragraph{Acknowledgement}
We thank the reviewers and the meta-reviewer for assessing our paper and for their constructive feedback. 
This work is supported by NSF grant RI:1939743.

\bibliography{bibfile,bib_spec}
\bibliographystyle{plainnat}
\input{supplement}

\end{document}

%% file: supplement.tex
\onecolumn
\appendix
\aistatstitle{Supplementary Materials}
\setcounter{section}{0}
\section{Conceptual Details}
\label{app:concept}

In this appendix section, we fill in more technical details and proofs from the main paper.

\subsection{Multi-class and Continuous Valued Fairness Risks}
\label{sec:multiclass_fairness}

Fairness metrics can be extended to \textbf{multi-class} labels and  sensitive features in a number of different ways \citep{rouzot2022learning,xian2023efficient}.
Here we recap the disparity of true positive rates (TPRs).
Let the label space be $[n_y] := \{1, \ldots, n_y\}$,
and the sensitive feature space be $[n_a]$.
Denote
\begin{align}
\nonumber
    \tpr(y, a) := P(\Yhat = y| Y = y, A = a), 
    \forall\ y \in [n_y], \ a \in [n_a].
\end{align}
Let $S_y$ be the set of labels of interest,
\eg, $S_y = [n_y]$.
Then we can define the fairness metric as
\begin{align}
    &\Delta_\tpr := \max_{a, a' \in [n_a]} \max_{y, y' \in S_y} 
    \abr{P(\Yhat = y| Y = y, A = a) 
        - P(\Yhat = y'| Y = y', A = a')}.
\end{align}
When $\Delta_\tpr = 0$ and the classes are binary, 
this fairness notion recovers equalized odds with $S_y = [n_y]$,
equal opportunity with $S_y = \{1\}$,
and predictive equality with $S_y = \{0\}$ \citep{Chouldechova2017Fair}.

Suppose there is a classifier $f$ that assigns the probability $P_f(\Yhat = y|x)$.
Then we can use the data to estimate 
\begin{align}
    \nonumber
    &P(\Yhat = y| Y = y, A = a) \approx
    \text{mean}\, S, \where 
    S := \{ P_f(\Yhat = y | x_i) : a_i = a, y_i = y \}.
\end{align}
Plugging it into $\Delta_\tpr$ would directly provide a differentiable fairness risk $\Fcal$ for regularization.

\paragraph{Demographic parity:}
we need $\Yhat$ to be independent of $A$,
and the degree of dependence can be measured by any existing applicable metric such as mutual information or 
\begin{align}
    &\max_{a, a' \in [n_a]} \max_{y \in S_y}
    \abr{P(\Yhat = y| A = a) 
        - P(\Yhat = y| A = a')} \\
    \text{or} \quad &
        \max_{a \in [n_a]} \max_{y \in S_y}
    \abr{P(\Yhat = y| A = a) - P(\Yhat = y)}
\end{align}
Again, we can estimate the relevant probabilities from data by
\begin{align}
\nonumber
    &P(\Yhat = y| A = a) \approx
    \text{mean}\, S,
    \where 
    S := \{ P_f(\Yhat=y|x_i) : a_i = a\} \\
    \nonumber
    &P(\Yhat = y) \approx
    \text{mean}\, S,
    \where 
    S := \{ P_f(\Yhat=y|x_i)\}.
\end{align}

\paragraph{Concrete fairness risk expressions}
For ease of reference, 
we explicitly write out the expressions.

For equalized odds
\begin{align}
\nonumber
    \Fcal_{\text{DEO}} &= 
    \max_{y \in \{0,1\}} |\text{mean} \{ P_f(\Yhat = y | x_i) : a_i = 0, y_i = y \} - \text{mean} \{ P_f(\Yhat = y | x_i) : a_i = 1, y_i = y \}| \\
\label{eq:def_deo}
    &=
    \max_{y \in \{0,1\}} |\text{mean} \{ P_f(\Yhat = 1 | x_i) : a_i = 0, y_i = y \} - \text{mean} \{ P_f(\Yhat = 1 | x_i) : a_i = 1, y_i = y \}|.    
\end{align}
For equal opportunity:
\begin{align}
\label{eq:def_deopp}
    \Fcal_{\text{DEOPP}} = 
    |\text{mean} \{ P_f(\Yhat = 1 | x_i) : a_i = 0, y_i = 1 \} 
    - \text{mean} \{ P_f(\Yhat = 1 | x_i) : a_i = 1, y_i = 1 \}|.
\end{align}
For demographic parity:
\begin{align}
\nonumber
    \Fcal_{\text{DDP}} &= 
\max_{y \in \{0,1\}} |\text{mean} \{ P_f(\Yhat = y | x_i) : a_i = 0\} - \text{mean} \{ P_f(\Yhat = y | x_i) : a_i = 1\}| \\
\label{eq:def_dpp}
    &= |\text{mean} \{ P_f(\Yhat = 1 | x_i) : a_i = 0\} - \text{mean} \{ P_f(\Yhat = 1 | x_i) : a_i = 1\}|.
\end{align} 

\newpage

\paragraph{Continuously valued label and sensitive features}
When the label and sensitive features are \textbf{continuous},
the above methods cease to be applicable.
We will resort to variational estimators such as InfoNCE,
and they require paired sampled of $\Yhat$ and $A$.
However, we can do better than that because the classifier $f$ provides a distribution of $\Yhat$ instead of a single sample.
For example, in mutual information neural estimation (MINE),
we can estimate the KL-divergence between $P_{A,\Yhat}$ and $P_{A} P_{\Yhat}$ by
\begin{align}
    \sup_{t: (A, \Yhat) \to \RR} 
    \bigg\{ \underbrace{\EE_{P_{A,\Yhat}} [t(A, \Yhat)]}_{\Circled{1}}
    - \underbrace{\EE_{P_A} \EE_{P_{\Yhat}} [\exp(t(A, \Yhat)-1)]}_{\Circled{2}} \bigg\}.
\end{align}
Then we can estimate both terms by
\begin{align}
    &\Circled{1} \approx \frac{1}{n} \sum_{i=1}^n \EE_{\Yhat \sim P_f(\Yhat|x_i)} [t(a_i, \Yhat)] \quad
    \Circled{2} \approx 
    \frac{1}{n^2} \sum_{i=1}^n \sum_{j=1}^n  \EE_{\Yhat \sim P_f(\Yhat|x_j)} [\exp(t(a_i, \Yhat)-1)].
\end{align}
If the function space of $t$ is simple enough,
we can directly evaluate the two expectations in a closed form.
Otherwise, we can sample $\Yhat$ from $P_f(\Yhat|x_j)$ and apply Gumbel-softmax for differentiation.

\subsection{Proof of Theorem \ref{thm:bound_sample}}
\label{sec:proof_sample_app}

\begin{proof}
    First we evaluate the bounded difference:
replace $Z_s$ with $Z'_s$ while keeping all other $Z_{s'}$ intact.
Here $Z'_s$ is any admissible assignment of $a^{(s)}_i$ and $y^{(s)}_i$. 
Since $\Fcal$ takes value in $[0,1]$,
it is trivial that $|\hat{\Ecal}_n(Z_1, \ldots, Z_N) - \hat{\Ecal}_n(Z_1,\ldots, Z_{s-1}, Z'_s, Z_{s+1}, \ldots, Z_N)| 
= \frac{1}{N} |\Fcal(P_f, Z_{s'}) - \Fcal(P_f, Z_s) | 
\le C/N$.
Further noting that  the expectation of $\hat{\Ecal}_n(Z_1, \ldots, Z_N)$ over $(Z_1,\ldots,Z_N)$ is $\Ecal$,
the McDiarmid's inequality immediately implies the theorem.    
\end{proof}

\begin{figure*}[t]
\begin{subfigure}{0.323\textwidth}
  \centering
  \includegraphics[width=\linewidth,clip=true, viewport = 0.3cm 0.1cm 16.2cm 14.1cm]{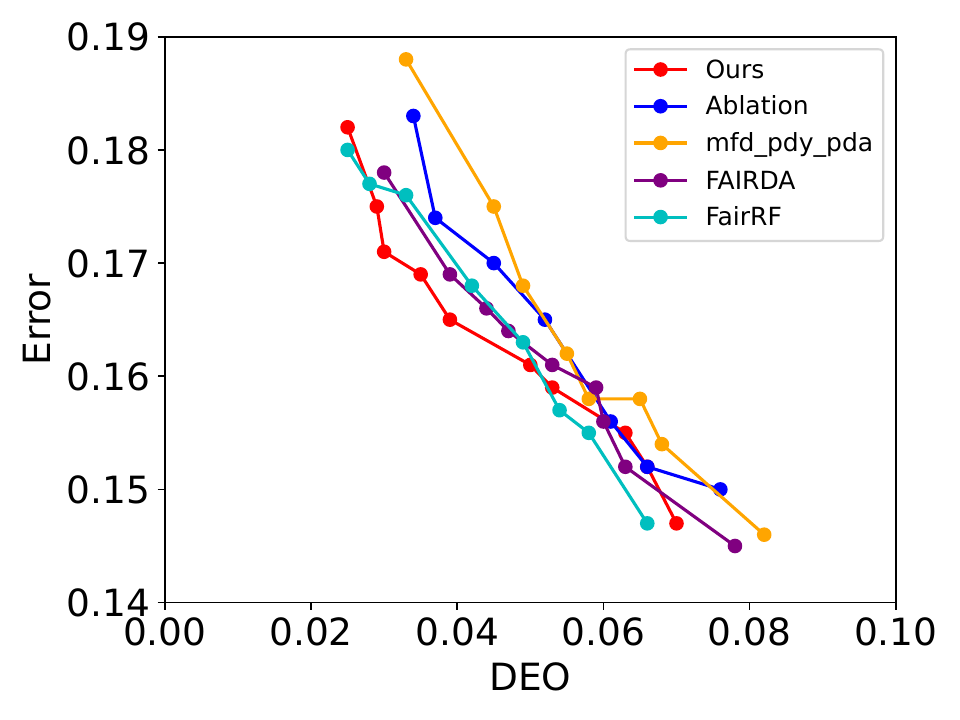}\quad
  \caption{DEO-Sparse}
\end{subfigure}
\begin{subfigure}{0.323\textwidth}
  \centering
  \includegraphics[width=\linewidth,clip=true, viewport = 0.3cm 0.1cm 16.2cm 14.1cm]{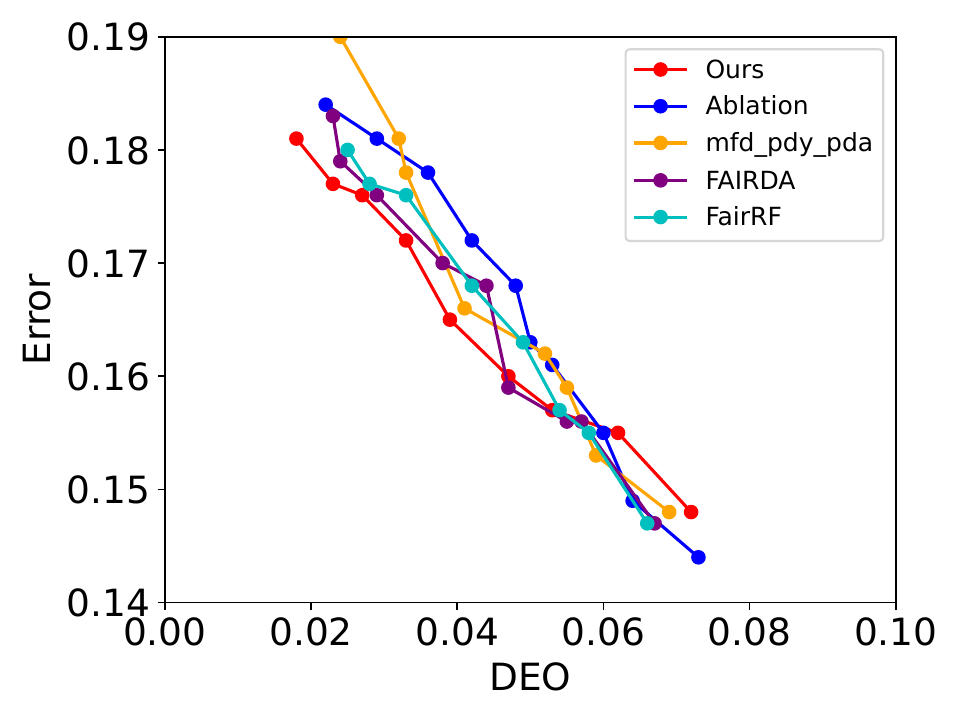}\quad
  \caption{DEO-Medium}
\end{subfigure}
\begin{subfigure}{0.323\textwidth}\quad
  \centering
  \includegraphics[width=\linewidth,clip=true, viewport = 0.3cm 0.1cm 16.2cm 14.1cm]{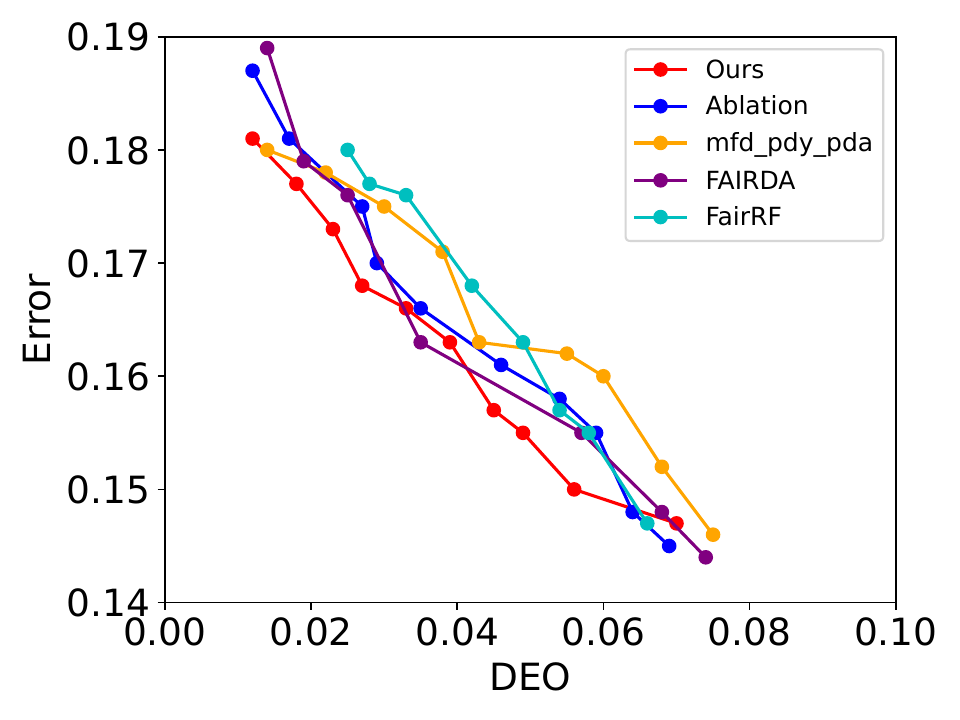}
  \caption{DEO-Dense}
\end{subfigure}
\\
  \centering
\begin{subfigure}{0.323\textwidth}
  \centering
  \includegraphics[width=\linewidth,clip=true, viewport = 0.3cm 0.1cm 16.2cm 14.1cm]{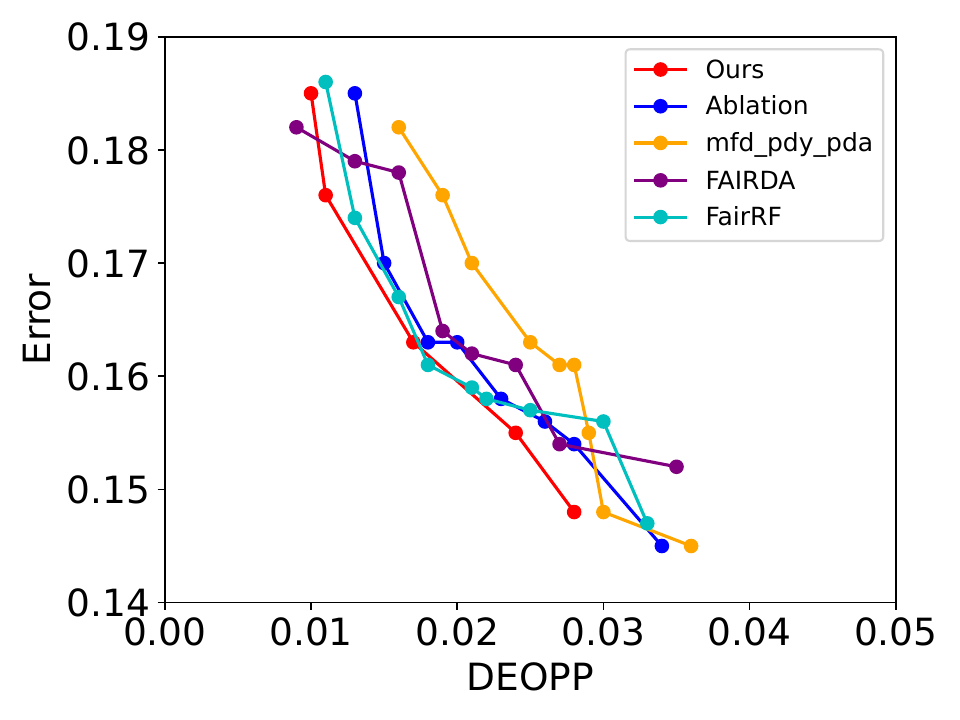}\quad
  \caption{DEOPP-Sparse}
\end{subfigure}
\begin{subfigure}{0.323\textwidth}
  \centering
  \includegraphics[width=\linewidth,clip=true, viewport = 0.3cm 0.1cm 16.2cm 14.1cm]{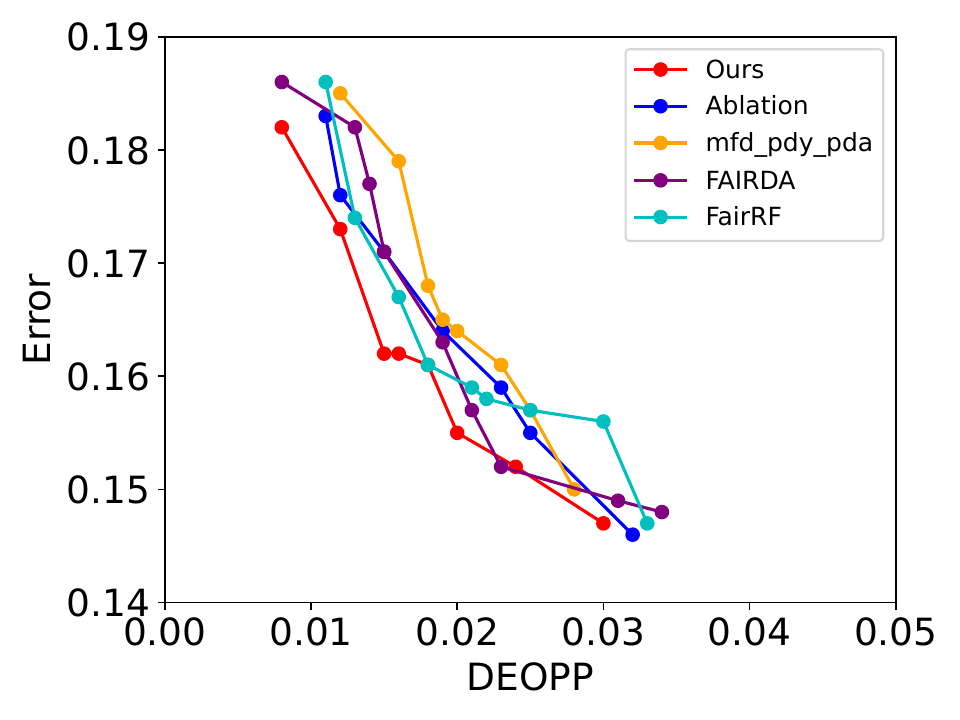}\quad
  \caption{DEOPP-Medium}
\end{subfigure}
\begin{subfigure}{0.323\textwidth}
  \centering
  \includegraphics[width=\linewidth,clip=true, viewport = 0.3cm 0.1cm 16.2cm 14.1cm]{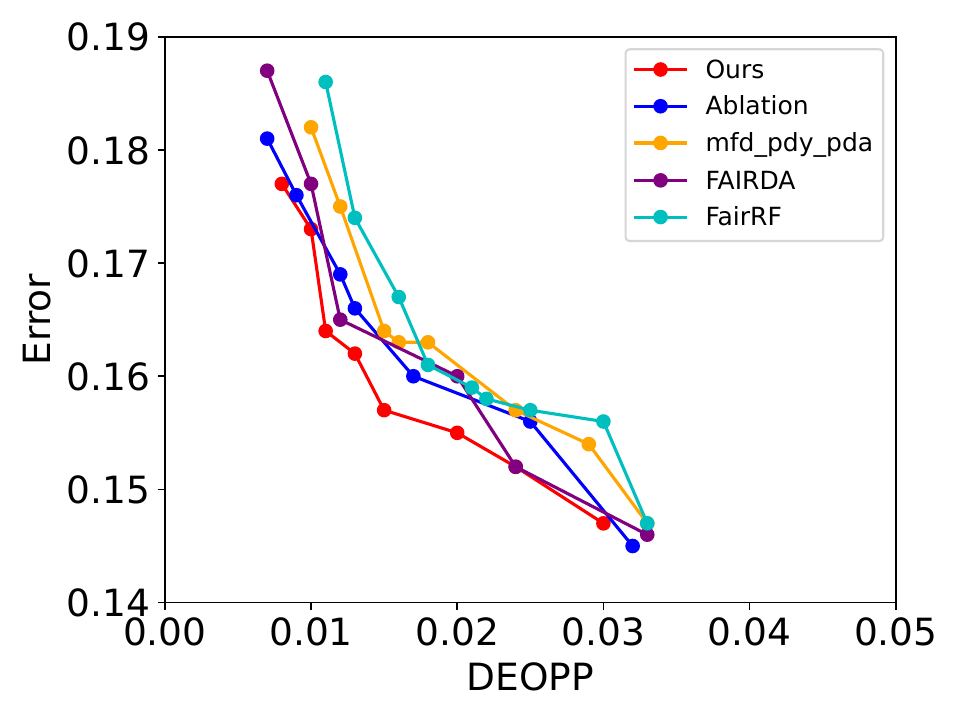}  
  \caption{DEOPP-Dense}
\end{subfigure}
\vspace{-0.3em}
\caption{Pareto frontier of error versus DEO/DEOPP for \textbf{Adult-Race}}
\label{fig:adult_race}
\end{figure*}

\subsection{Straight-through Gumbel-Softmax for vanilla fairness risk}
\label{sec:gumble_app}

Suppose a Bernoulli variable $y_i$ has $P(y_i = 1) = q_i$.
Then the vanilla Gumbel-Softmax method first draws i.i.d. samples from a Gumbel$(0,1)$ distribution:
\begin{align}
    \{\alpha^{(s)}_i, \beta^{(s)}_i : i = 1,\ldots,n, \text{ and } \ s = 1,\ldots,N\}.
\end{align} 
Then a sample of $y^{(s)}_i$ is constructed by
\begin{align}
\label{eq:ys_gumbel}
    y^{(s)}_i =  \frac{\exp(\frac{\log q_i + \alpha^{(s)}_i}{T})}{\exp( \frac{\log q_i + \alpha^{(s)}_i}{T}) + \exp(\frac{\log (1 - q_i) + \beta^{(s)}_i}{T})} \ ,
\end{align}
where $T > 0$ is a temperature parameter.
A larger value of $T$ leads to a more smooth and uniform sample, 
i.e., $y^{(s)}_i$ will be closer to 0.5.
Clearly $y^{(s)}_i$ is not discrete,
but for reasonably small $T$, it will be close to 0 or 1.
To patch up the non-integrality of $y^{(s)}_i$, 
the straight-through Gumbel-Softmax gradient estimator only uses the differentiable variable in the backward gradient propagation, 
while the forward pass still uses categorical variables 
(\ie, turning $y^{(s)}_i$ into 1 if it is above 0.5, and 0 otherwise).
The same approach can be applied to sample $a_i$.

\subsection{ELBO}
\label{sec:app_ELBO}

We can extend the ELBO as
\begin{align*}
    \log p_{\theta}(\xvec,\atil, \ytil) 
    &\geq 
    \!\!\expunder{(\zvec, a, y) \sim q_{\phi}(\cdot|\xvec,\atil,\ytil)}
    [\log p_\theta(\xvec, \atil, \ytil|a, y, \zvec) 
    \! +\log (p(y)  p(a) p(\zvec))
    - \log q_\phi(a, y, \zvec| \xvec, \atil, \ytil)] \\ 
    &= 
    \expunder{\zvec \sim q_{\phi}(\zvec|\xvec)} \log p_\theta(x|a, y, \zvec) 
    + \expunder{y \sim q_\phi(y|\xvec, \ytil)} \log P_\theta(\ytil|y) + \expunder{a \sim q_\phi(a|\xvec, \atil)} \log P_\theta(\atil|a)  \\  
    &- \text{KL}[ q_\phi(\zvec|\xvec) || p(\zvec)] 
    - \text{KL}[ q_\phi(y|\xvec, \ytil) || p(y)]
    - \text{KL}[ q_\phi(a|\xvec, \atil) || p(a)]
    \\
    &=: \text{ELBO}(\xvec,\atil, \ytil).
\end{align*}

\subsection{Soften the step function to enable differentiation}

The step function $\sembrack{\cdot > 0}$ offers no useful derivative.
Noting that taking a threshold of a probability $f(X)$ is equivalent to thresholding its logit $\sigma^{-1}(f(X))$,
we improve numerical performance by setting 
$P_f(Y=1|X) = \sigma(\frac{\sigma^{-1}(f(X)) - \tau}{T})$,
where the temperature $T$ controls the steepness of the approximation.
Finally, group-specific threshold can be enabled by replacing $\tau$ with $\tau_1 a_i + \tau_0 (1-a_i)$.
Given the values of $a_i$---either from the training data or from the $N$ samples---it is straightforward to optimize this objective.


\subsection{Labeling test data by minimizing fairness and classification risks}
\label{sec:inf_risk_app}

With the learned $\phi$, 
the test data can be labeled by simply seeking the $\yvec$ that 
maximizes the likelihood while also minimizing the expected fairness risk:
\begin{align}
\label{test_objective}
\nonumber
    &\argmin_{y_i \in \{0,1\}} \Bigl\{ -\frac{1}{n_{\text{test}}} \sum_{i} \log q_\phi(y_i|\xvec_i) 
    + \expunder{A_i \sim {\color{red} q_\phi}(A | \xvec_i, \atil_i)} \Fcal(P_{q_\phi(y|\xvec)}, \yvec, A_{1:n_{\text{test}}}) \Bigr\}.
\end{align}
When all demographics are observed, 
this can be solved by tuning group-specific thresholds on $q_\phi(y_i = 1|\xvec_i)$.
When test demographics are missing (possibly in their entirety),
we can first compute the binary demographics via $\argmax_A q_\phi(A|\xvec_i, \atil_i)$,
and then tune the group-specific threshold on $q_\phi(y_i = 1|\xvec_i)$.
Afterwards, we do a coordinate descent to finetune $\yvec$ in the face of full expectation on $A_i$ for $a_i = \emptyset$ using Monte Carlo.

\subsection{Extension to class-probability estimation}
\label{sec:ext_class_prob_app}

The above discussion has been intentionally kept general by considering a randomized classifier for $Y|X$.
In practice, however, this is not quite realistic because one does not get to predict multiple times for a single test set.
Therefore, we would like to specialize the framework to classification based on class-probability,
\ie, $P_f(Y=1|x) = \sembrack{f(x) - \tau > 0}$.
It also provides the convenience of tuning the parameter $\tau$.
Once $h$ is learned and fixed thereafter, 
different test scenarios may demand different trade-offs between risks in classification and fairness.
For example, \citet{Jang2022Group} proposed adapting the thresholds for each demographic group.
To enable differentiation, the $\sembrack{\cdot}$ operator can be softened as shown in Appendix~\ref{app:concept}.

\section{Experiment Settings}
We implemented our algorithm using Python 3.11.5 and PyTorch 2.1.1. The code runs on a server with Ubuntu 18.04, powered by an Intel(R) Core(TM) i9-9900X CPU @ 3.50GHz and 128 GB of RAM. For GPU resources, we have four GeForce RTX 2080 Ti cards, each with 11 GB of graphic memory, and the GPU driver version is 455.23.05.
\section{Additional Experimental Results}
\label{sec:app_experiment}

\begin{figure*}[t]
\begin{subfigure}{0.323\textwidth}
  \centering
  \includegraphics[width=\linewidth,clip=true, viewport = -0.1cm 0.1cm 16.4cm 14.1cm]{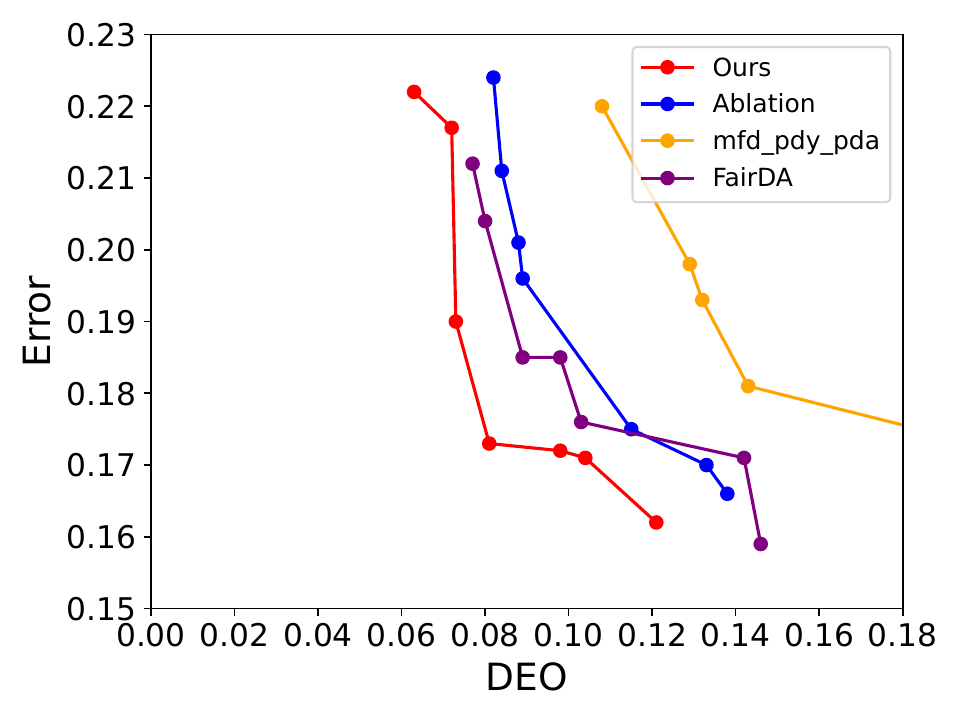}\quad
  \caption{DEO-Sparse}
\end{subfigure}
\begin{subfigure}{0.323\textwidth}
  \centering
  \includegraphics[width=\linewidth,clip=true, viewport = -0.1cm 0.1cm 16.4cm 14.1cm]{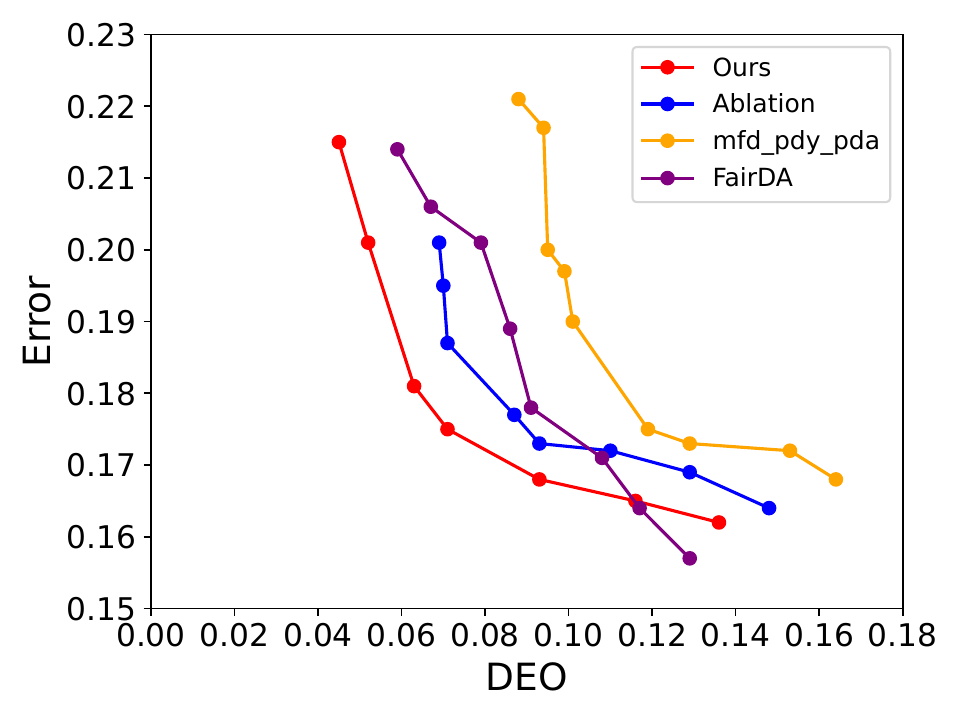}\quad
  \caption{DEO-Medium}
\end{subfigure}
\begin{subfigure}{0.323\textwidth}\quad
  \centering
  \includegraphics[width=\linewidth,clip=true, viewport = -0.1cm 0.1cm 16.4cm 14.1cm]{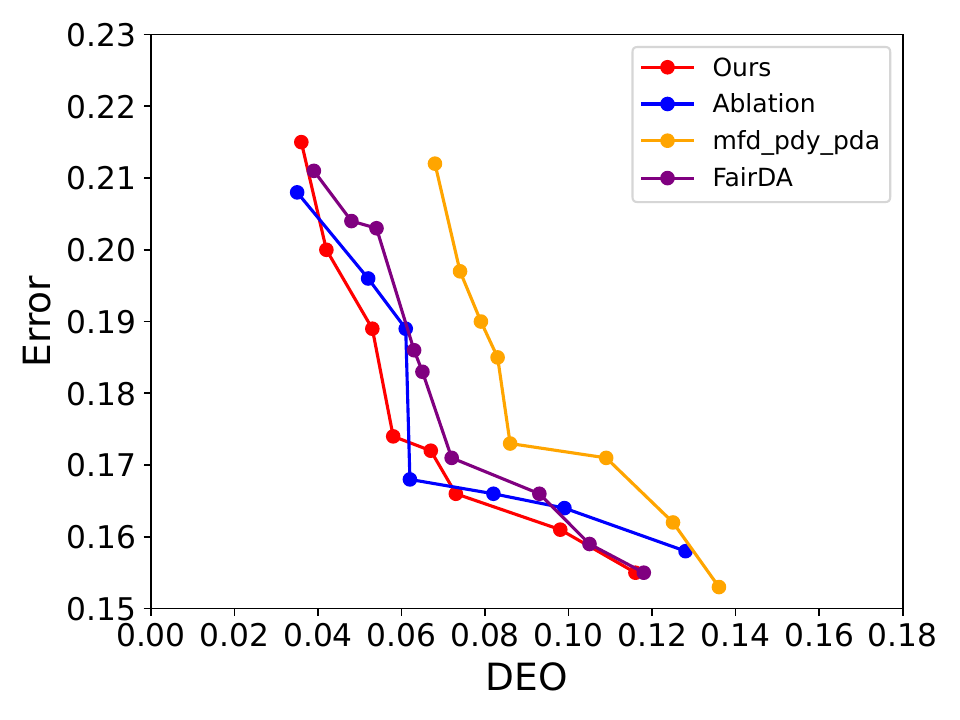}
  \caption{DEO-Dense}
\end{subfigure}
\\
  \centering
\begin{subfigure}{0.323\textwidth}
  \centering
  \includegraphics[width=\linewidth,clip=true, viewport = -0.1cm 0.1cm 16.4cm 14.1cm]{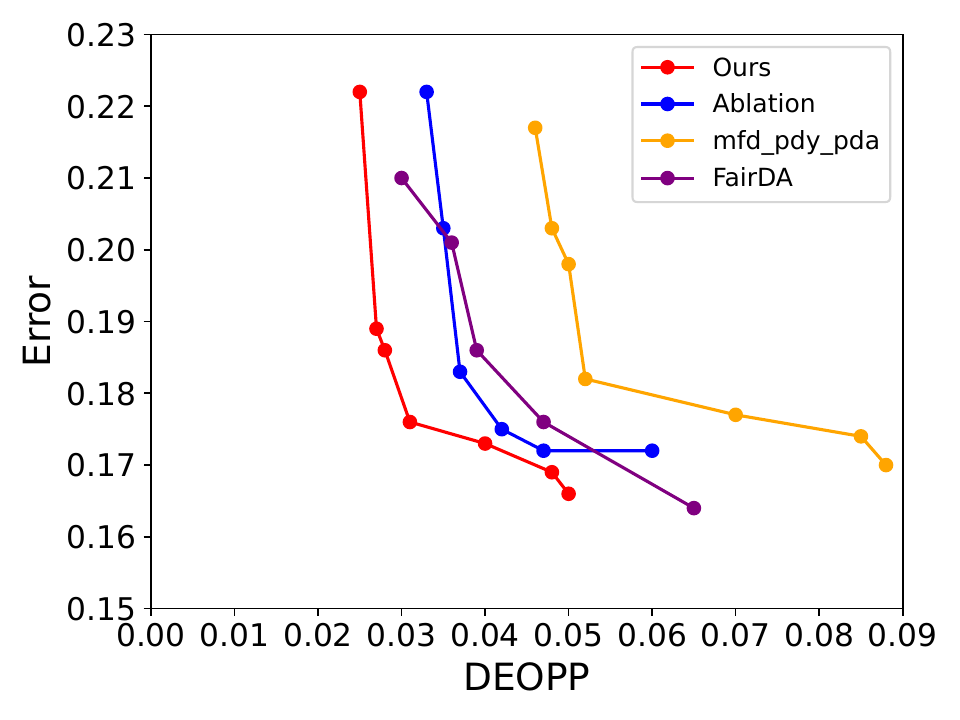}\quad
  \caption{DEOPP-Sparse}
\end{subfigure}
\begin{subfigure}{0.323\textwidth}
  \centering
  \includegraphics[width=\linewidth,clip=true, viewport = -0.1cm 0.1cm 16.4cm 14.1cm]{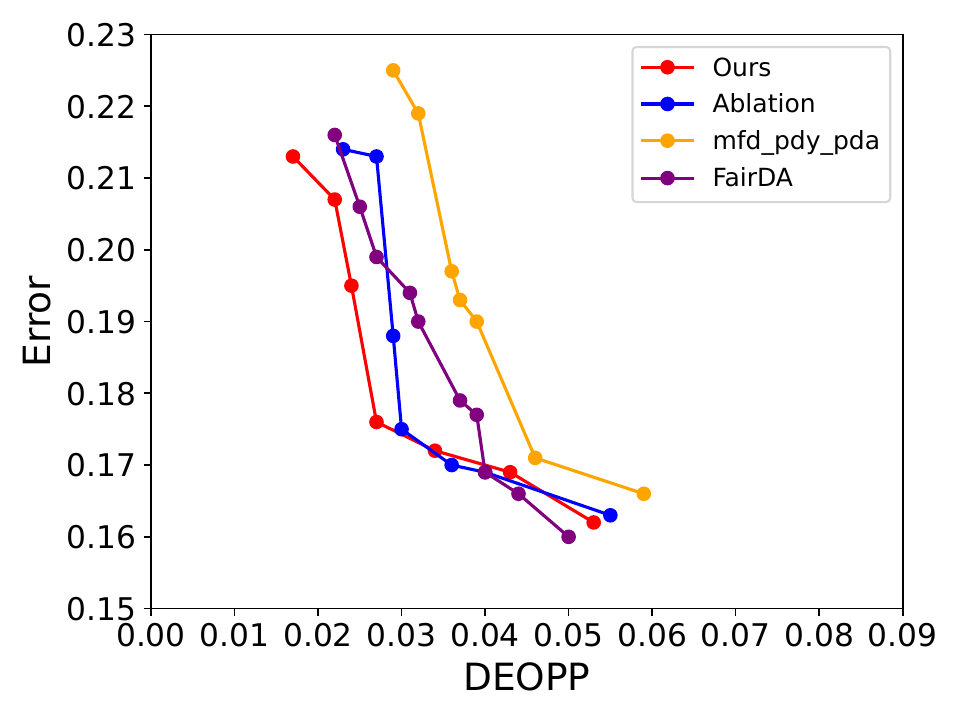}\quad
  \caption{DEOPP-Medium}
\end{subfigure}
\begin{subfigure}{0.323\textwidth}
  \centering
  \includegraphics[width=\linewidth,clip=true, viewport = -0.1cm 0.1cm 16.4cm 14.1cm]{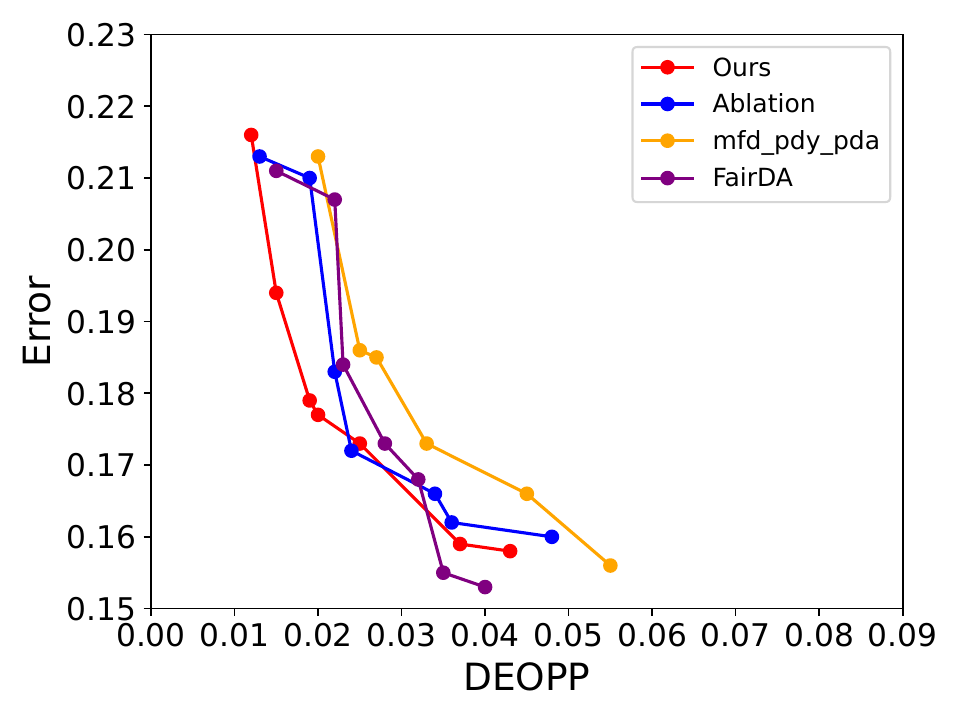}
  \caption{DEOPP-Dense}
\end{subfigure}
\vspace{-0.3em}
\caption{Pareto frontier of error versus DEO/DEOPP for \textbf{CelebA-HighCheekbones}}
\label{fig:celeba_highcheekbones}
\vspace{-0.3em}
\end{figure*}

Figure~\ref{fig:adult_race} shows the Pareto frontier of error versus DEO/DEOPP for Adult-Race. \highlight{Figure~\ref{fig:celeba_highcheekbones} shows the results for Celeba-HighCheekbones. Here, we use high cheekbones as the label and gender as the group attribute.}

\begin{figure*}[h]
\begin{subfigure}{0.323\textwidth}
  \centering
  \includegraphics[width=\linewidth,clip=true, viewport = 0.2cm 0.1cm 16.2cm 14.1cm]{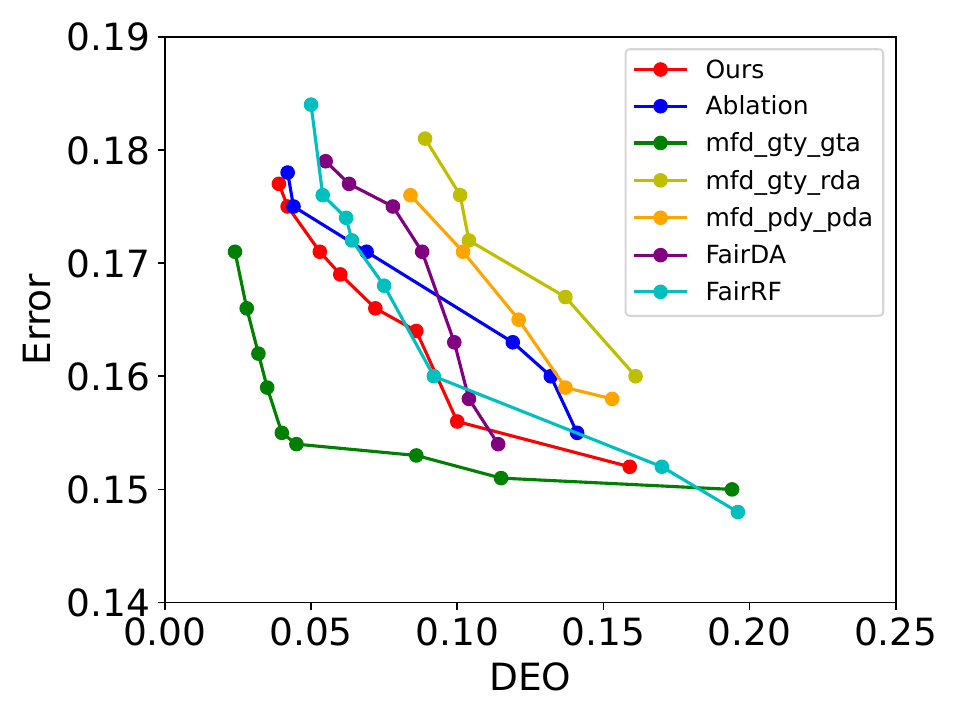}\quad
  \caption{DEO-Sparse}
\end{subfigure}
\begin{subfigure}{0.323\textwidth}
  \centering
  \includegraphics[width=\linewidth,clip=true, viewport = 0.2cm 0.1cm 16.2cm 14.1cm]{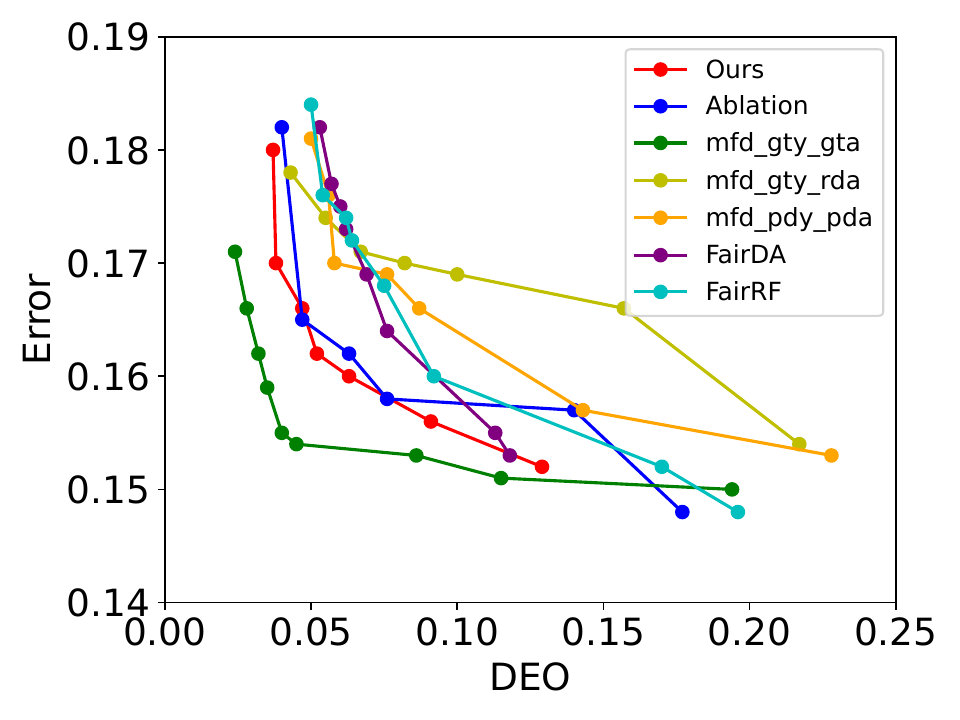}\quad
  \caption{DEO-Medium}
\end{subfigure}
\begin{subfigure}{0.323\textwidth}\quad
  \centering
  \includegraphics[width=\linewidth,clip=true, viewport = 0.2cm 0.1cm 16.2cm 14.1cm]{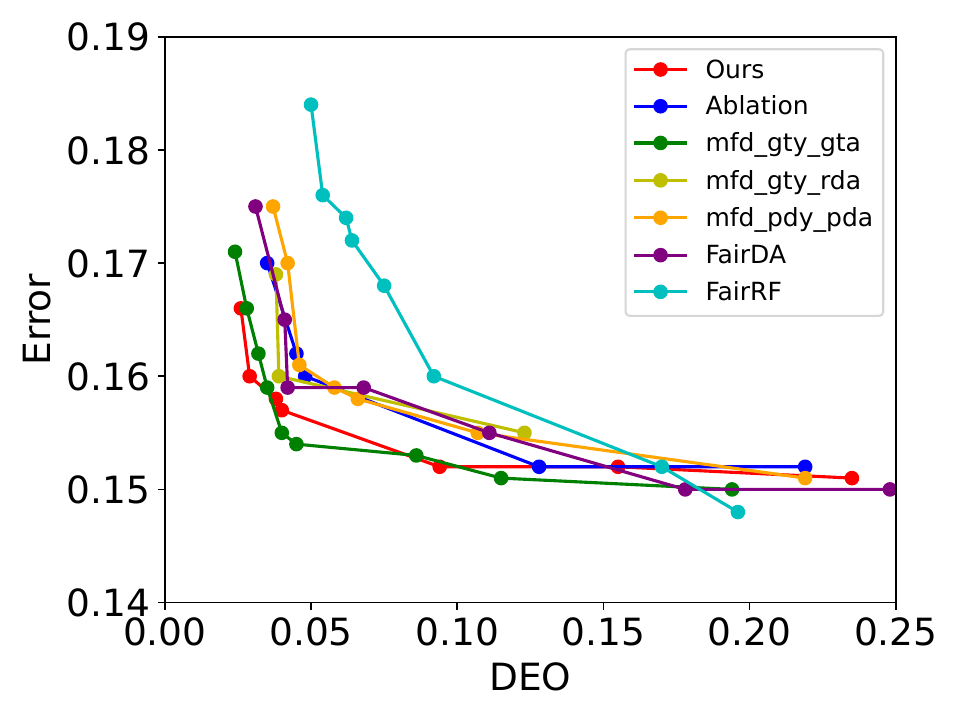}
  \caption{DEO-Dense}
\end{subfigure}
\\
  \centering
\begin{subfigure}{0.323\textwidth}
  \centering
  \includegraphics[width=\linewidth,clip=true, viewport = 0.3cm 0.1cm 17.1cm 14.1cm]{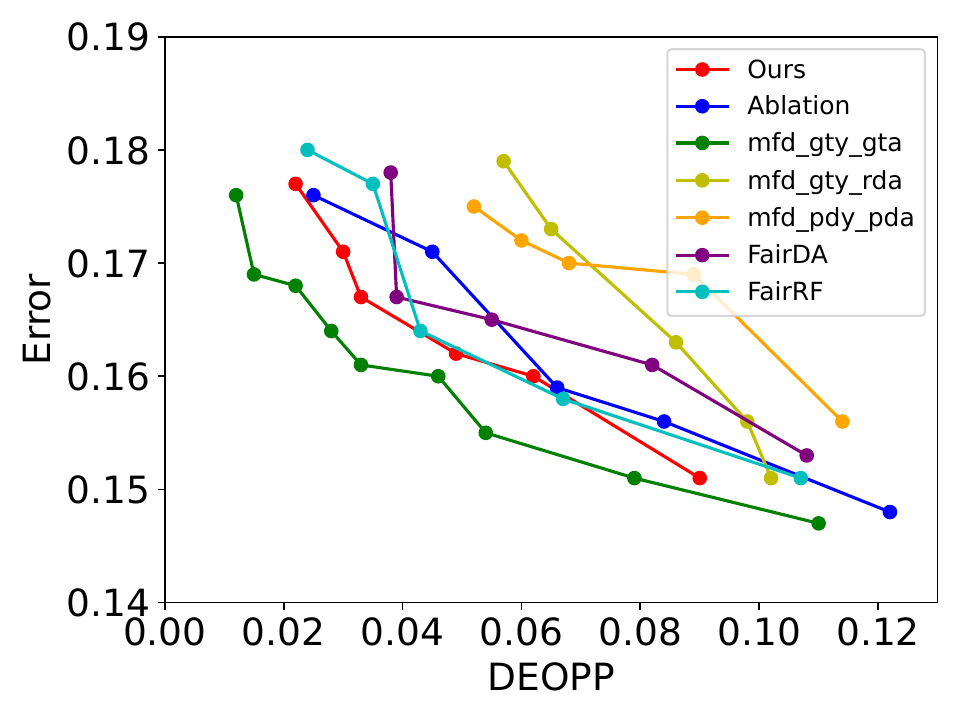}\quad
  \caption{DEOPP-Sparse}
\end{subfigure}
\begin{subfigure}{0.323\textwidth}
  \centering
  \includegraphics[width=\linewidth,clip=true, viewport = 0.3cm 0.1cm 17.1cm 14.1cm]{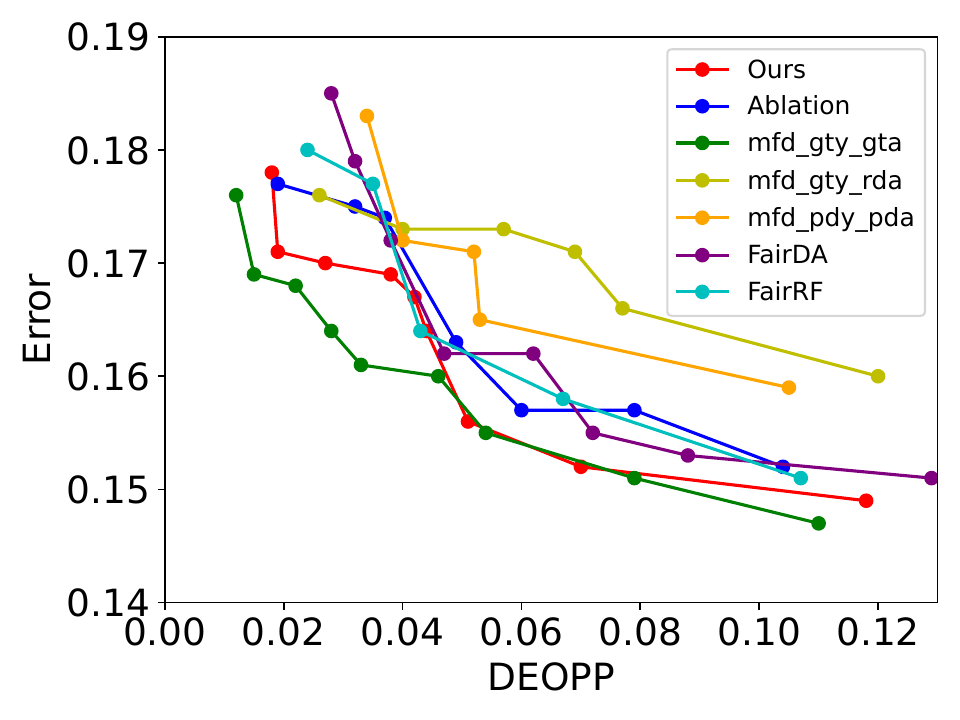}\quad
  \caption{DEOPP-Medium}
\end{subfigure}
\begin{subfigure}{0.323\textwidth}
  \centering
  \includegraphics[width=\linewidth,clip=true, viewport = 0.3cm 0.1cm 17.1cm 14.1cm]{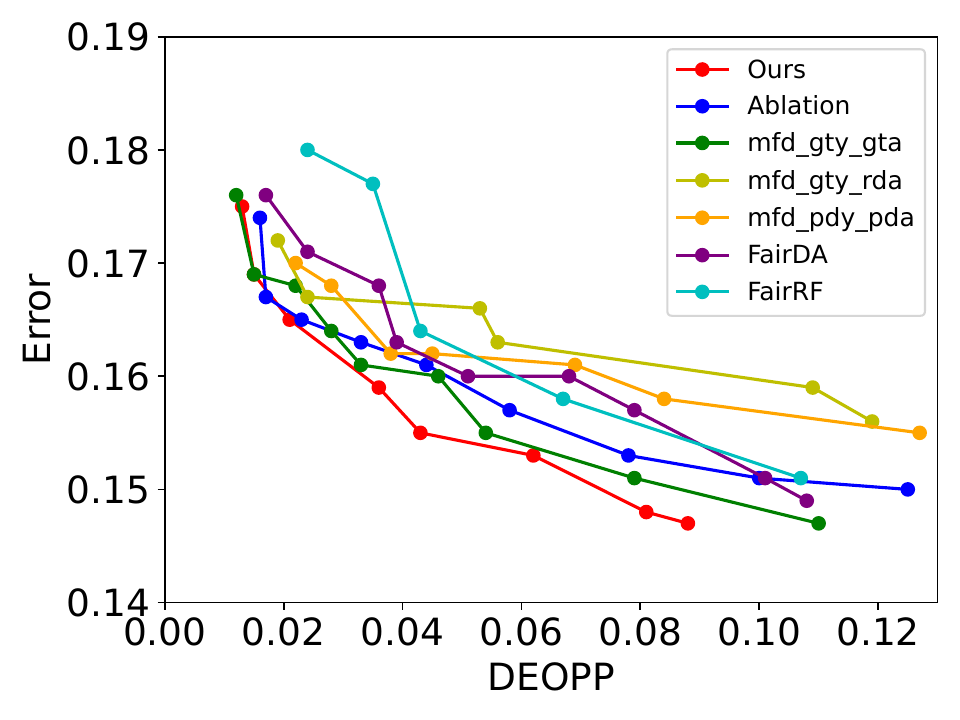}
  \caption{DEOPP-Dense}
\end{subfigure}
\caption{Pareto frontier of error versus DEO/DEOPP for Adult-Gender}
\label{fig:adult_gender_appendix}
\end{figure*}

\begin{figure*}[h]
\begin{subfigure}{0.323\textwidth}
  \centering
  \includegraphics[width=\linewidth,clip=true, viewport = 0.2cm 0.1cm 16.3cm 14.1cm]{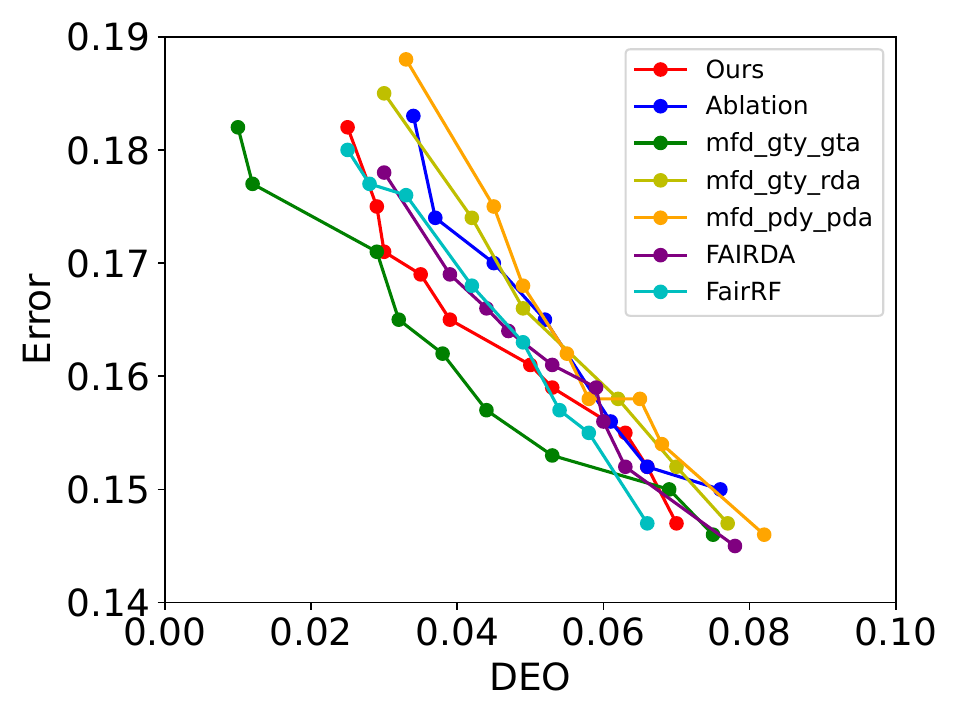}\quad
  \caption{DEO-Sparse}
\end{subfigure}
\begin{subfigure}{0.323\textwidth}
  \centering
  \includegraphics[width=\linewidth,clip=true, viewport = 0.2cm 0.1cm 16.3cm 14.1cm]{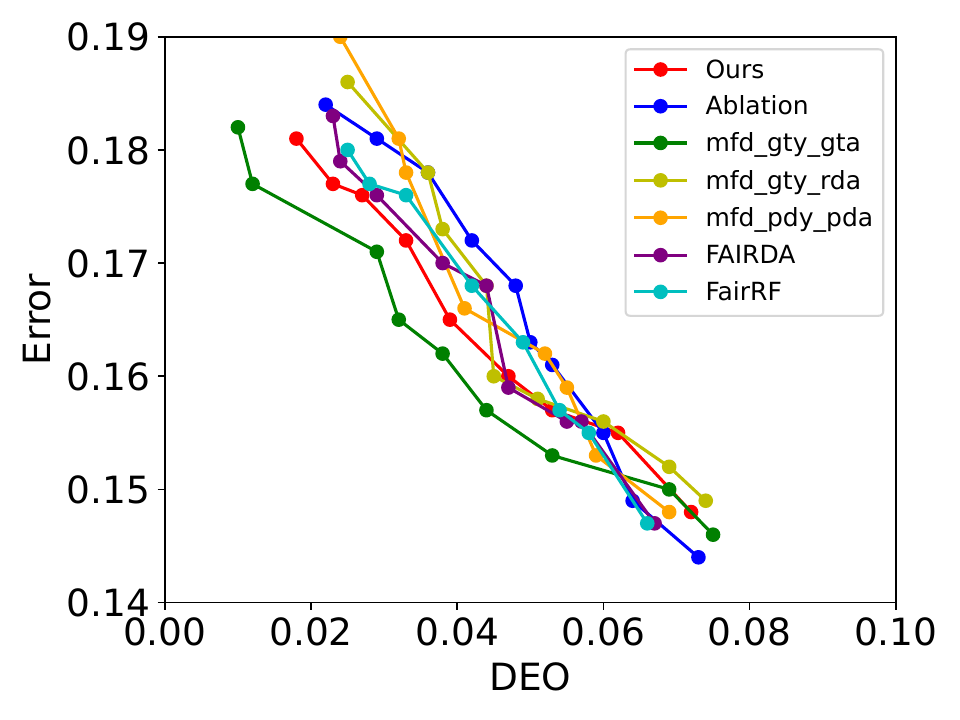}\quad
  \caption{DEO-Medium}
\end{subfigure}
\begin{subfigure}{0.323\textwidth}\quad
  \centering
  \includegraphics[width=\linewidth,clip=true, viewport = 0.2cm 0.1cm 16.3cm 14.1cm]{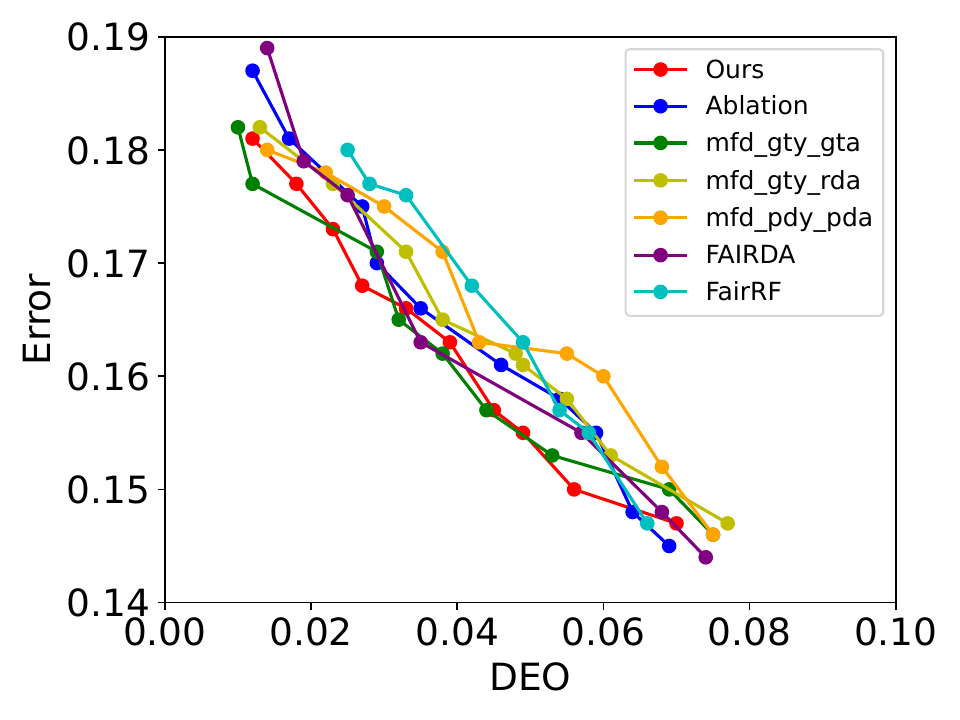}
  \caption{DEO-Dense}
\end{subfigure}
\\
  \centering
\begin{subfigure}{0.323\textwidth}
  \centering
  \includegraphics[width=\linewidth,clip=true, viewport = 0.3cm 0.1cm 16.4cm 14.1cm]{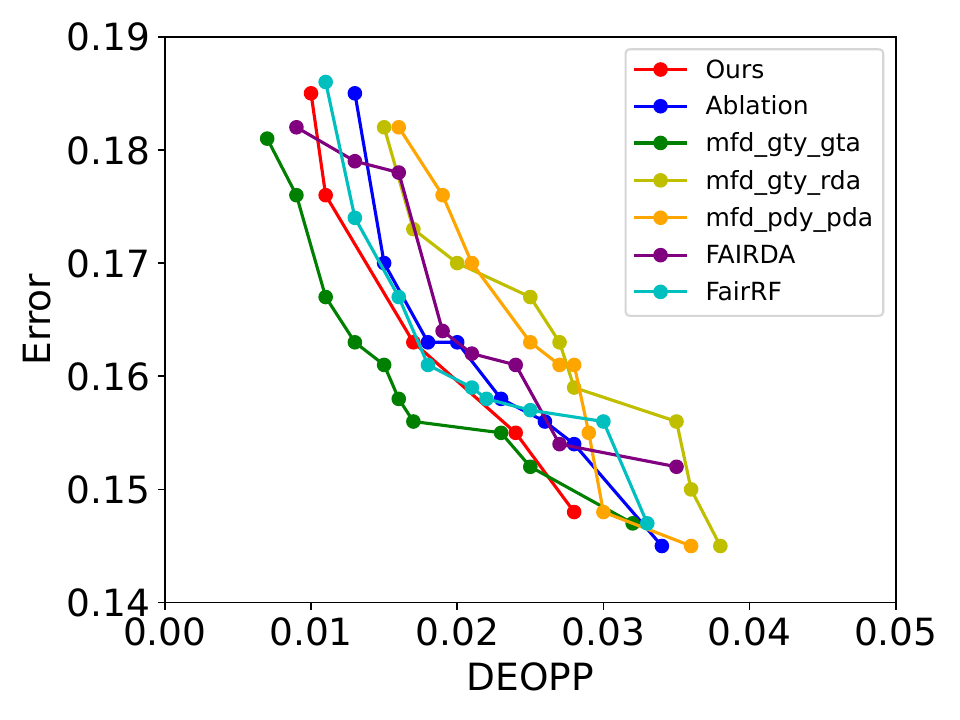}\quad
  \caption{DEOPP-Sparse}
\end{subfigure}
\begin{subfigure}{0.323\textwidth}
  \centering
  \includegraphics[width=\linewidth,clip=true, viewport = 0.3cm 0.1cm 16.4cm 14.1cm]{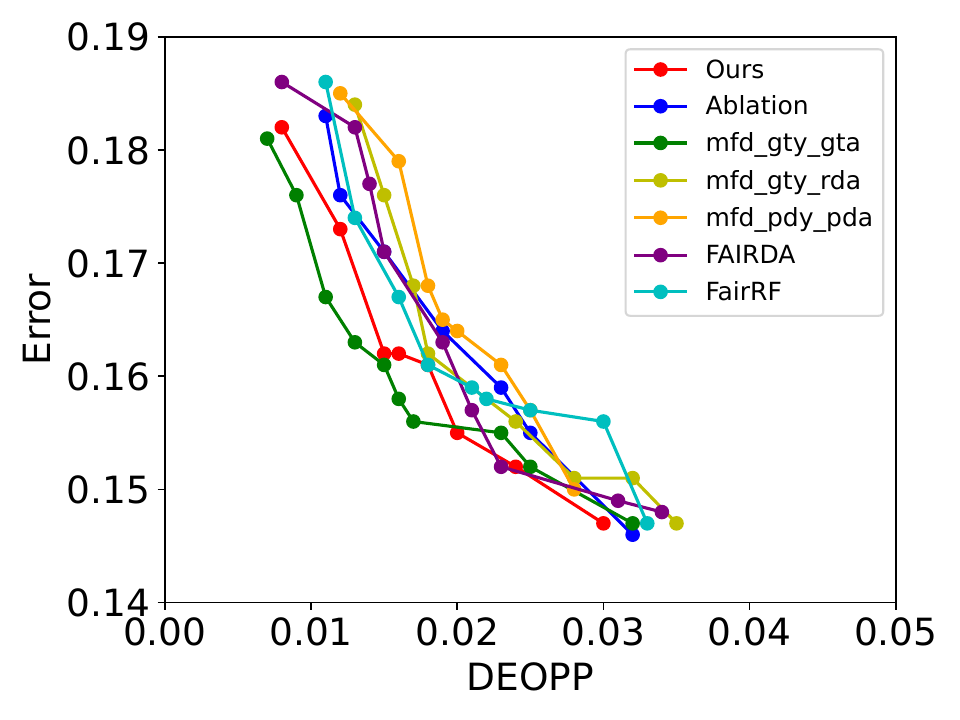}\quad
  \caption{DEOPP-Medium}
\end{subfigure}
\begin{subfigure}{0.323\textwidth}
  \centering
  \includegraphics[width=\linewidth,clip=true, viewport = 0.3cm 0.1cm 16.4cm 14.1cm]{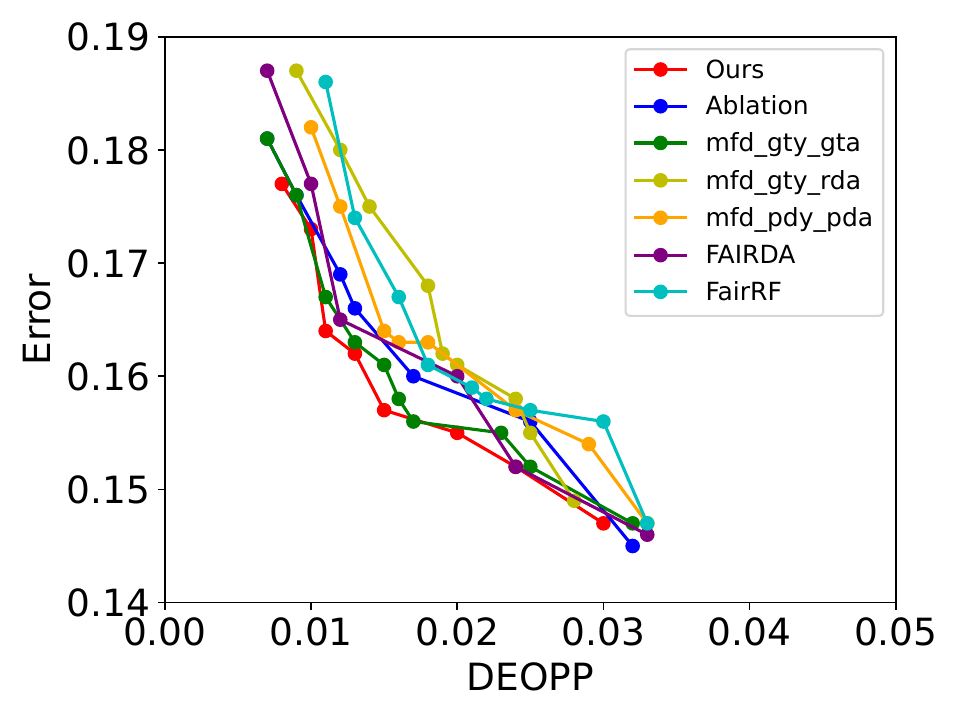}
  \caption{DEOPP-Dense}
\end{subfigure}
\caption{Pareto frontier of error versus DEO/DEOPP for Adult-Race}
\label{fig:adult_race_appendix}
\end{figure*}

\begin{figure*}[h]
\begin{subfigure}{0.323\textwidth}
  \centering
  \includegraphics[width=\linewidth,clip=true, viewport = -0.1cm 0.1cm 16.4cm 14.1cm]{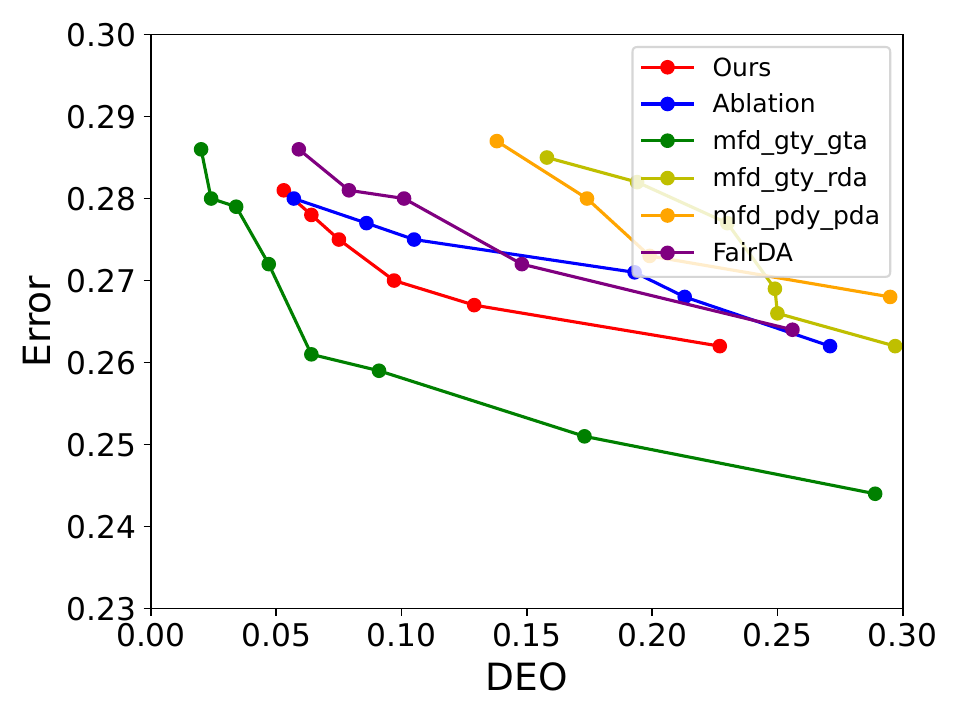}\quad
  \caption{DEO-Sparse}
\end{subfigure}
\begin{subfigure}{0.323\textwidth}
  \centering
  \includegraphics[width=\linewidth,clip=true, viewport = -0.1cm 0.1cm 16.4cm 14.1cm]{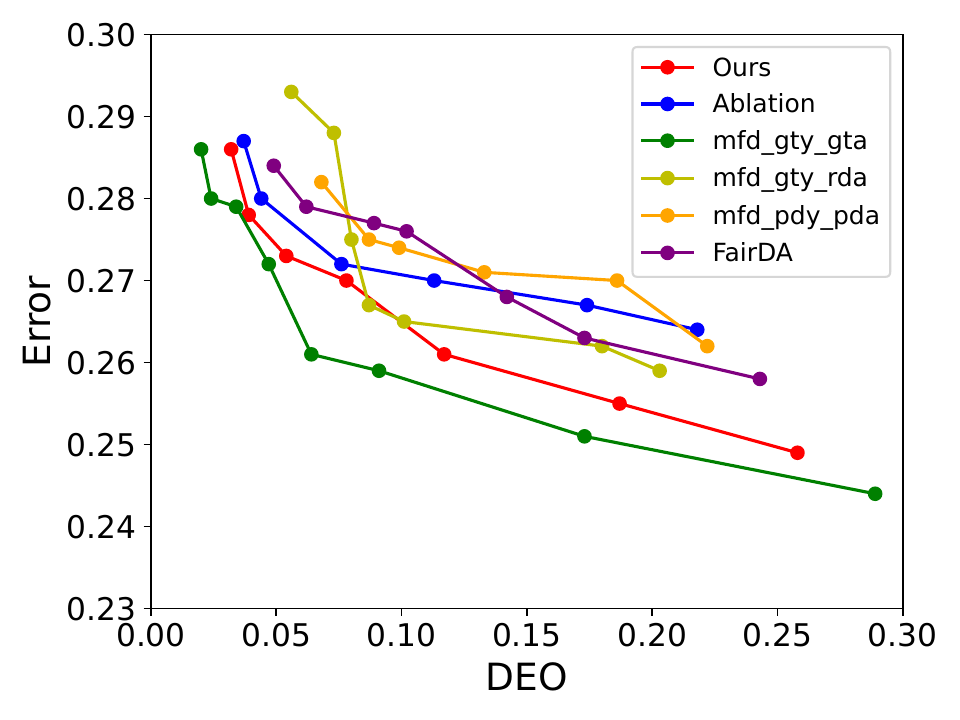}\quad
  \caption{DEO-Medium}
\end{subfigure}
\begin{subfigure}{0.323\textwidth}\quad
  \centering
  \includegraphics[width=\linewidth,clip=true, viewport = -0.1cm 0.1cm 16.4cm 14.1cm]{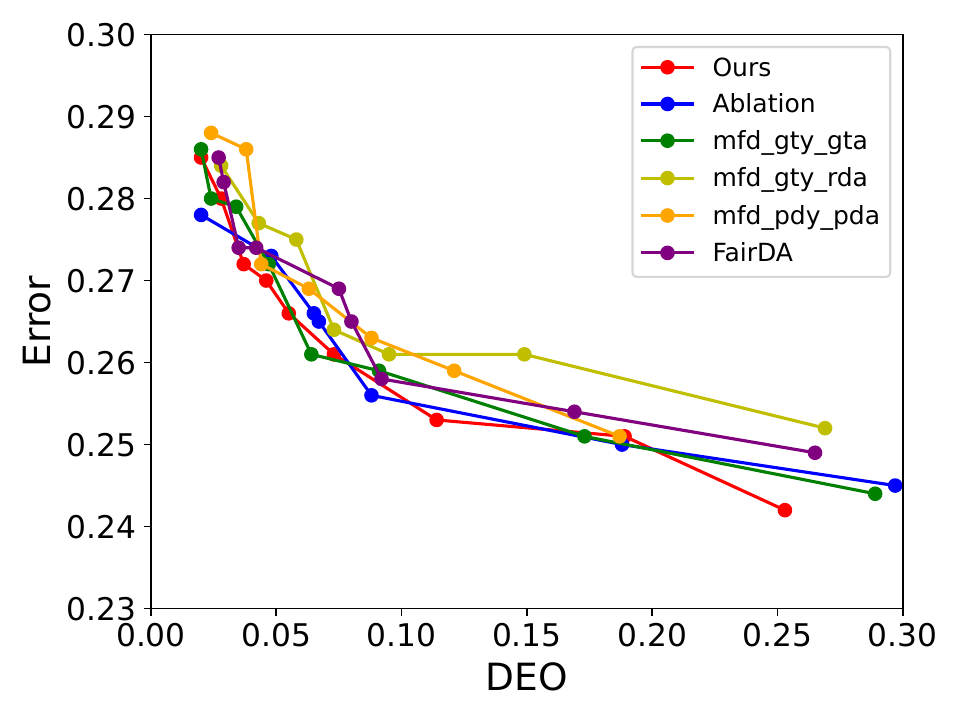}
  \caption{DEO-Dense}
\end{subfigure}
\\
  \centering
\begin{subfigure}{0.323\textwidth}
  \centering
  \includegraphics[width=\linewidth,clip=true, viewport = 0.0cm 0.1cm 17.3cm 14.1cm]{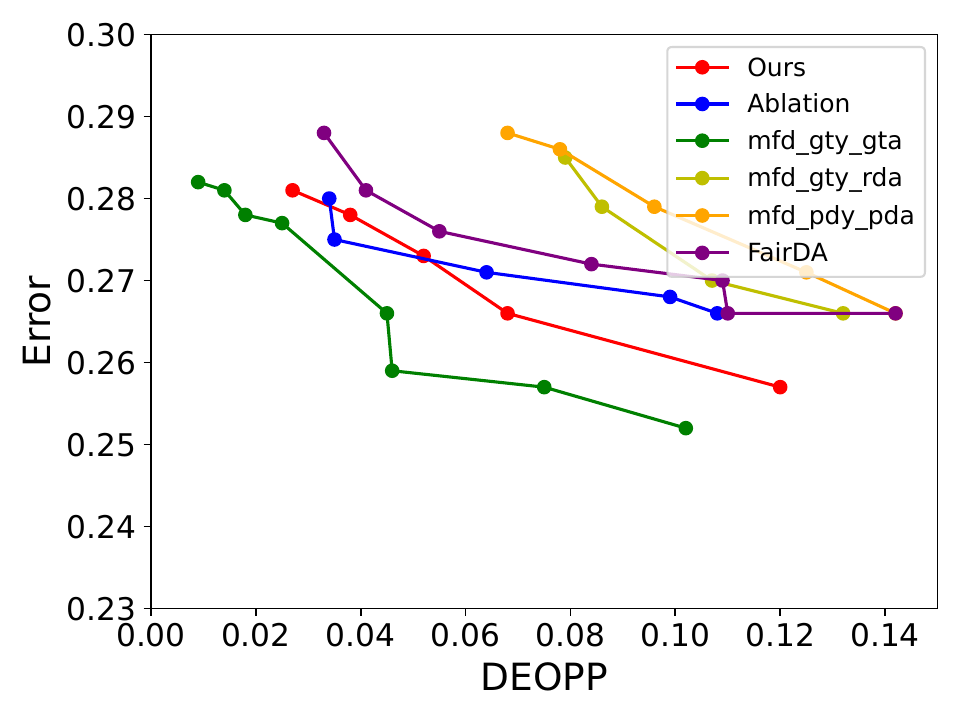}\quad
  \caption{DEOPP-Sparse}
\end{subfigure}
\begin{subfigure}{0.323\textwidth}
  \centering
  \includegraphics[width=\linewidth,clip=true, viewport = 0.0cm 0.1cm 17.3cm 14.1cm]{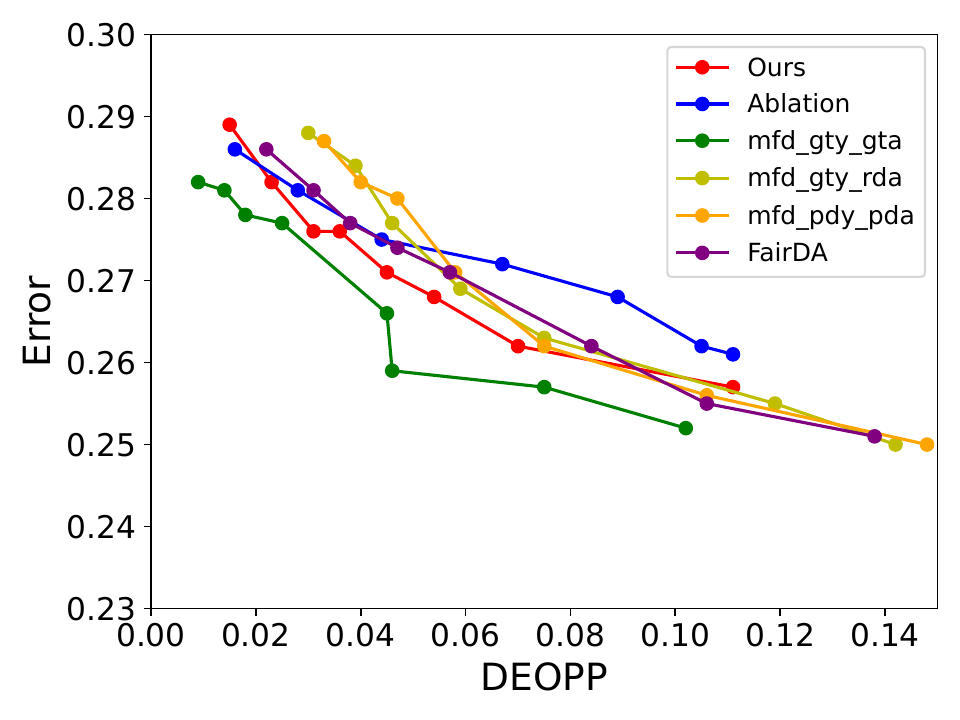}\quad
  \caption{DEOPP-Medium}
\end{subfigure}
\begin{subfigure}{0.323\textwidth}
  \centering
  \includegraphics[width=\linewidth,clip=true, viewport = 0.0cm 0.1cm 17.3cm 14.1cm]{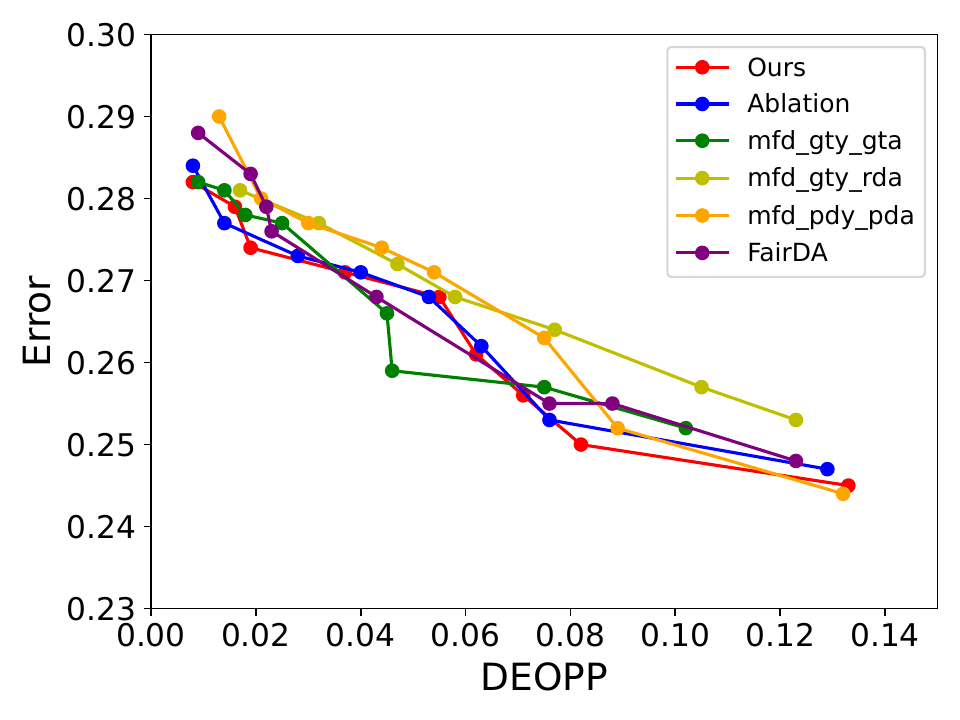}
  \caption{DEOPP-Dense}
\end{subfigure}
\caption{Pareto frontier of error versus DEO/DEOPP for CelebA}
\label{fig:celeba_appendix}
\end{figure*}

We additionally present the Pareto frontier attained by all the baseline methods. 
Figures~\ref{fig:adult_gender_appendix} to \ref{fig:celeba_appendix} are crowded, 
hence moved to the appendix.
The curve of mfd\_gty\_gta remains intact across all sparsity levels for a given dataset,
because it imputes the sensitive feature with its ground truth. 
%
Note that it is not fair to compare Fair-SS-VAE against mfd\_gty\_gta, 
because the latter has full access to the ground truth of both labels and sensitive features. 
Nonetheless, we still visualize its Pareto frontier because it serves as a rough upper bound for all the methods.

\section{Source Code}

\textbf{The experiment code is available at
\url{https://tinyurl.com/4wndtfbb}.}

\clearpage


\section{Statistical Property of Estimator \eqref{eq:mean_est_DP}}
\label{sec:property_estimator}

\begin{theorem}
    The following estimator of $\theta := P(\Yhat = 1 | A = a)$ is consistent and \textbf{asymptotically} unbiased:
    \begin{align}
        \theta_n := \sum\nolimits_{i: a_i = a} P(\Yhat = 1 | x_i) \Big / \sum\nolimits_{i: a_i = a} 1 \, .
    \end{align}  
    However, it is in general not unbiased.
\end{theorem}

\begin{proof}
    To show consistency, note
    \begin{align}
        \frac{1}{n} \sum\nolimits_{i: a_i = a} 1
        \ &\xrightarrow{P} \ 
        P(A = a) \\
        \frac{1}{n} \sum \nolimits_{i: a_i = a} P(\Yhat = 1 | x_i)
        \ &\xrightarrow{P} \
        P(\Yhat = 1, A = a).
    \end{align}
    Therefore,
    \begin{align}
        \frac{\sum\nolimits_{i: a_i = a} P(\Yhat = 1 | x_i) }{\sum\nolimits_{i: a_i = a} 1}
        = 
        \frac{\frac{1}{n} \sum \nolimits_{i: a_i = a} P(\Yhat = 1 | x_i)}{\frac{1}{n} \sum\nolimits_{i: a_i = a} 1}
        \ \xrightarrow{P} \
        \frac{P(\Yhat = 1, A = a)}{P(A = a)}
        = P(\Yhat = 1 | A = a).
    \end{align}   

    To show asymptotic unbiasedness,
    note
    \begin{align}
        \EE[|\theta_n - \theta|]
        &= \EE [|\theta_n - \theta| \cdot 
        \sembrack{|\theta_n - \theta| < \epsilon}]
        + \EE [|\theta_n - \theta| 
        \cdot \sembrack{|\theta_n - \theta| \ge \epsilon}] \\
         &\le \epsilon + \EE [|\theta_n - \theta| 
        \cdot \sembrack{|\theta_n - \theta| \ge \epsilon}]
        \\
        \text{(by Cauchy-Schwarz)} \quad &\le \epsilon + \sqrt{\EE [|\theta_n - \theta|^2] \Pr ( |\theta_n - \theta| \ge \epsilon )}.
    \end{align}
    Consistency has already established that $\Pr ( |\theta_n - \theta| \ge \epsilon ) \to 0$ as $n \to \infty$.
    Since $\theta_n$ is clearly bounded by 1,
    the right-hand side falls below $2\epsilon$ for sufficiently large $n$.
\end{proof}

\vfill

\newpage

%% file: paper.bbl
\begin{thebibliography}{52}
\providecommand{\natexlab}[1]{#1}
\providecommand{\url}[1]{\texttt{#1}}
\expandafter\ifx\csname urlstyle\endcsname\relax
  \providecommand{\doi}[1]{doi: #1}\else
  \providecommand{\doi}{doi: \begingroup \urlstyle{rm}\Url}\fi

\bibitem[Barocas et~al.(2019)Barocas, Hardt, and Narayanan]{barocas2019fairness}
Solon Barocas, Moritz Hardt, and Arvind Narayanan.
\newblock \emph{Fairness and Machine Learning: Limitations and Opportunities}.
\newblock fairmlbook.org, 2019.
\newblock \url{http://www.fairmlbook.org}.

\bibitem[Becker and Kohavi(1996)]{misc_adult_2}
Barry Becker and Ronny Kohavi.
\newblock {Adult}.
\newblock UCI Machine Learning Repository, 1996.
\newblock {DOI}: https://doi.org/10.24432/C5XW20.

\bibitem[Bellamy et~al.(2018)Bellamy, Dey, Hind, Hoffman, Houde, Kannan, Lohia, Martino, Mehta, Mojsilovic, et~al.]{bellamy2018ai}
Rachel~KE Bellamy, Kuntal Dey, Michael Hind, Samuel~C Hoffman, Stephanie Houde, Kalapriya Kannan, Pranay Lohia, Jacquelyn Martino, Sameep Mehta, Aleksandra Mojsilovic, et~al.
\newblock {AI} fairness 360: An extensible toolkit for detecting, understanding, and mitigating unwanted algorithmic bias.
\newblock \emph{arXiv preprint arXiv:1810.01943}, 2018.

\bibitem[Buolamwini and Gebru(2018)]{buolamwini2018gender}
Joy Buolamwini and Timnit Gebru.
\newblock Gender shades: Intersectional accuracy disparities in commercial gender classification.
\newblock In \emph{Proceedings of the 1st Conference on Fairness, Accountability and Transparency}, 2018.

\bibitem[Calders and Verwer(2010)]{calders2010three}
Toon Calders and Sicco Verwer.
\newblock Three naive bayes approaches for discrimination-free classification.
\newblock \emph{Data Mining and Knowledge Discovery}, 21\penalty0 (2):\penalty0 277–292, sep 2010.

\bibitem[Celis et~al.(2021{\natexlab{a}})Celis, Huang, Keswani, and Vishnoi]{celis2021fair}
L~Elisa Celis, Lingxiao Huang, Vijay Keswani, and Nisheeth~K Vishnoi.
\newblock Fair classification with noisy protected attributes: A framework with provable guarantees.
\newblock In \emph{International Conference on Machine Learning (ICML)}, 2021{\natexlab{a}}.

\bibitem[Celis et~al.(2021{\natexlab{b}})Celis, Mehrotra, and Vishnoi]{celis2021fairb}
L~Elisa Celis, Anay Mehrotra, and Nisheeth Vishnoi.
\newblock Fair classification with adversarial perturbations.
\newblock In \emph{Advances in Neural Information Processing Systems (NeurIPS)}, 2021{\natexlab{b}}.

\bibitem[Chen et~al.(2019)Chen, Kallus, Mao, Svacha, and Udell]{chen2019fairness}
Jiahao Chen, Nathan Kallus, Xiaojie Mao, Geoffry Svacha, and Madeleine Udell.
\newblock Fairness under unawareness: Assessing disparity when protected class is unobserved.
\newblock In \emph{Proceedings of the Conference on Fairness, Accountability, and Transparency}, 2019.

\bibitem[Cho et~al.(2020)Cho, Hwang, and Suh]{Cho2020fair}
Jaewoong Cho, Gyeongjo Hwang, and Changho Suh.
\newblock A fair classifier using kernel density estimation.
\newblock In \emph{Advances in Neural Information Processing Systems (NeurIPS)}, 2020.

\bibitem[Chouldechova(2017)]{Chouldechova2017Fair}
Alexandra Chouldechova.
\newblock Fair prediction with disparate impact: A study of bias in recidivism prediction instruments.
\newblock \emph{Big Data}, 5\penalty0 (2):\penalty0 153--163, 2017.

\bibitem[{Consumer Financial Protection Bureau}(2023)]{Consumer23}
{Consumer Financial Protection Bureau}.
\newblock {12 CFR Part 1002 - Equal Credit Opportunity Act (Regulation B), 1002.5 Rules concerning requests for information}.
\newblock 2023.

\bibitem[Coston et~al.(2019)Coston, Ramamurthy, Wei, Varshney, Speakman, Mustahsan, and Chakraborty]{coston2019fair}
Amanda Coston, Karthikeyan~Natesan Ramamurthy, Dennis Wei, Kush~R. Varshney, Skyler Speakman, Zairah Mustahsan, and Supriyo Chakraborty.
\newblock Fair transfer learning with missing protected attributes.
\newblock In \emph{Proceedings of the 2019 AAAI/ACM Conference on AI, Ethics, and Society}, 2019.

\bibitem[Dai and Wang(2021)]{DaiWan21}
Enyan Dai and Suhang Wang.
\newblock Say no to the discrimination: Learning fair graph neural networks with limited sensitive attribute information.
\newblock In \emph{International Conference on Web Search and Data Mining (WSDM)}, 2021.

\bibitem[Dwork et~al.(2012)Dwork, Hardt, Pitassi, Reingold, and Zemel]{dwork2012fairness}
Cynthia Dwork, Moritz Hardt, Toniann Pitassi, Omer Reingold, and Richard Zemel.
\newblock Fairness through awareness.
\newblock In \emph{Proceedings of the 3rd Innovations in Theoretical Computer Science Conference}, 2012.

\bibitem[Feldman et~al.(2015)Feldman, Friedler, Moeller, Scheidegger, and Venkatasubramanian]{feldman2015certifying}
Michael Feldman, Sorelle~A. Friedler, John Moeller, Carlos Scheidegger, and Suresh Venkatasubramanian.
\newblock Certifying and removing disparate impact.
\newblock In \emph{ACM SIGKDD Conference on Knowledge Discovery and Data Mining (KDD)}, 2015.

\bibitem[Gianfrancesco et~al.(2018)Gianfrancesco, Tamang, Yazdany, and Schmajuk]{gianfrancesco2018potential}
Milena~A. Gianfrancesco, Suzanne Tamang, Jinoos Yazdany, and Gabriela Schmajuk.
\newblock {Potential Biases in Machine Learning Algorithms Using Electronic Health Record Data}.
\newblock \emph{JAMA Internal Medicine}, 178\penalty0 (11):\penalty0 1544--1547, 2018.

\bibitem[Gupta et~al.(2018)Gupta, Cotter, Fard, and Wang]{gupta2018proxy}
Maya Gupta, Andrew Cotter, Mahdi~Milani Fard, and Serena Wang.
\newblock Proxy fairness.
\newblock \emph{arXiv:1806.11212}, 2018.

\bibitem[Hardt et~al.(2016)Hardt, Price, and Srebro]{hardt2016equality}
Moritz Hardt, Eric Price, and Nati Srebro.
\newblock Equality of opportunity in supervised learning.
\newblock In \emph{Advances in Neural Information Processing Systems (NeurIPS)}, pages 3315--3323, 2016.

\bibitem[Hashimoto et~al.(2018)Hashimoto, Srivastava, Namkoong, and Liang]{hashimoto2018Fairness}
Tatsunori Hashimoto, Megha Srivastava, Hongseok Namkoong, and Percy Liang.
\newblock Fairness without demographics in repeated loss minimization.
\newblock In \emph{International Conference on Machine Learning (ICML)}, 2018.

\bibitem[Jang et~al.(2017)Jang, Gu, and Poole]{Jang2017categorical}
Eric Jang, Shixiang Gu, and Ben Poole.
\newblock Categorical reparametrization with gumbel-softmax.
\newblock In \emph{International Conference on Learning Representations (ICLR)}, 2017.

\bibitem[Jang et~al.(2022)Jang, Shi, and Wang]{Jang2022Group}
Taeuk Jang, Pengyi Shi, and Xiaoqian Wang.
\newblock Group-aware threshold adaptation for fair classification.
\newblock In \emph{National Conference of Artificial Intelligence (AAAI)}, 2022.

\bibitem[Jung et~al.(2021)Jung, Lee, Park, and Moon]{jung2021fair}
Sangwon Jung, Donggyu Lee, Taeeon Park, and Taesup Moon.
\newblock Fair feature distillation for visual recognition.
\newblock In \emph{IEEE Conference on Computer Vision and Pattern Recognition (CVPR)}, pages 12115--12124, 2021.

\bibitem[Jung et~al.(2022)Jung, Chun, and Moon]{jung2022learning}
Sangwon Jung, Sanghyuk Chun, and Taesup Moon.
\newblock Learning fair classifiers with partially annotated group labels.
\newblock In \emph{IEEE Conference on Computer Vision and Pattern Recognition (CVPR)}, pages 10348--10357, June 2022.

\bibitem[Kallus et~al.(2022)Kallus, Mao, and Zhou]{kallus2022assessing}
Nathan Kallus, Xiaojie Mao, and Angela Zhou.
\newblock Assessing algorithmic fairness with unobserved protected class using data combination.
\newblock \emph{Management Science}, 68\penalty0 (3):\penalty0 1959--1981, 2022.

\bibitem[Kamiran and Calders(2012)]{kamiran2012data}
F.~Kamiran and T.G.K. Calders.
\newblock Data preprocessing techniques for classification without discrimination.
\newblock \emph{Knowledge and Information Systems}, 33\penalty0 (1):\penalty0 1--33, 2012.

\bibitem[Kingma et~al.(2014)Kingma, Mohamed, Rezende, and Welling]{Kingma2014semi}
Diederik~P Kingma, Shakir Mohamed, Danilo~Jimenez Rezende, and Max Welling.
\newblock Semi-supervised learning with deep generative models.
\newblock In \emph{Advances in Neural Information Processing Systems (NeurIPS)}, 2014.

\bibitem[Krumpal(2013)]{krumpal2013determinants}
Ivar Krumpal.
\newblock Determinants of social desirability bias in sensitive surveys: a literature review.
\newblock \emph{Quality \& quantity}, 47\penalty0 (4):\penalty0 2025--2047, 2013.

\bibitem[Lahoti et~al.(2020)Lahoti, Beutel, Chen, Lee, Prost, Thain, Wang, and Chi]{lahoti2020fairness}
Preethi Lahoti, Alex Beutel, Jilin Chen, Kang Lee, Flavien Prost, Nithum Thain, Xuezhi Wang, and Ed~Chi.
\newblock Fairness without demographics through adversarially reweighted learning.
\newblock In \emph{Advances in Neural Information Processing Systems (NeurIPS)}, 2020.

\bibitem[Lamy et~al.(2019)Lamy, Zhong, Menon, and Verma]{lamy2019noise}
Alexandre Lamy, Ziyuan Zhong, Aditya~Krishna Menon, and Nakul Verma.
\newblock Noise-tolerant fair classification.
\newblock In \emph{Advances in Neural Information Processing Systems (NeurIPS)}, 2019.

\bibitem[Liang et~al.(2023)Liang, Chen, Tian, and Shu]{liang2023fair}
Yueqing Liang, Canyu Chen, Tian Tian, and Kai Shu.
\newblock Fair classification via domain adaptation: A dual adversarial learning approach.
\newblock \emph{Frontiers in Big Data}, 5:\penalty0 129, 2023.

\bibitem[Lipton et~al.(2018)Lipton, McAuley, and Chouldechova]{lipton2018does}
Zachary Lipton, Julian McAuley, and Alexandra Chouldechova.
\newblock Does mitigating ml\textquotesingle s impact disparity require treatment disparity?
\newblock In \emph{Advances in Neural Information Processing Systems (NeurIPS)}, 2018.

\bibitem[Liu et~al.(2015)Liu, Luo, Wang, and Tang]{liu2015deep}
Ziwei Liu, Ping Luo, Xiaogang Wang, and Xiaoou Tang.
\newblock Deep learning face attributes in the wild.
\newblock In \emph{International Conference on Computer Vision (ICCV)}, pages 3730--3738, 2015.

\bibitem[Maddison et~al.(2017)Maddison, Mnih, and Teh]{maddison2017the}
Chris~J. Maddison, Andriy Mnih, and Yee~Whye Teh.
\newblock The concrete distribution: A continuous relaxation of discrete random variables.
\newblock In \emph{International Conference on Learning Representations (ICLR)}, 2017.

\bibitem[Mary et~al.(2019)Mary, Calauzenes, and El~Karoui]{mary2019fairness}
J{\'e}r{\'e}mie Mary, Cl{\'e}ment Calauzenes, and Noureddine El~Karoui.
\newblock Fairness-aware learning for continuous attributes and treatments.
\newblock In \emph{International Conference on Machine Learning (ICML)}, pages 4382--4391. PMLR, 2019.

\bibitem[Mehrabi et~al.(2021)Mehrabi, Morstatter, Saxena, Lerman, and Galstyan]{mehrabi2021survey}
Ninareh Mehrabi, Fred Morstatter, Nripsuta Saxena, Kristina Lerman, and Aram Galstyan.
\newblock A survey on bias and fairness in machine learning.
\newblock \emph{ACM Comput. Surv.}, 54\penalty0 (6), 2021.

\bibitem[Menon and Williamson(2018)]{menon2018cost}
Aditya~Krishna Menon and Robert~C. Williamson.
\newblock The cost of fairness in binary classification.
\newblock In \emph{Proceedings of the 1st Conference on Fairness, Accountability and Transparency}, 2018.

\bibitem[Mozannar et~al.(2020)Mozannar, Ohannessian, and Srebro]{mozannar2020fair}
Hussein Mozannar, Mesrob Ohannessian, and Nathan Srebro.
\newblock Fair learning with private demographic data.
\newblock In \emph{International Conference on Machine Learning (ICML)}, 2020.

\bibitem[Narayanaswamy et~al.(2017)Narayanaswamy, Paige, Van~de Meent, Desmaison, Goodman, Kohli, Wood, and Torr]{Narayanaswamy2017learning}
Siddharth Narayanaswamy, Timothy Paige, Jan-Willem Van~de Meent, Alban Desmaison, Noah Goodman, Pushmeet Kohli, Frank Wood, and Philip Torr.
\newblock Learning disentangled representations with semi-supervised deep generative models.
\newblock In \emph{Advances in Neural Information Processing Systems (NeurIPS)}, 2017.

\bibitem[Rouzot et~al.(2022)Rouzot, Ferry, and Huguet]{rouzot2022learning}
Julien Rouzot, Julien Ferry, and Marie-Jos{\'e} Huguet.
\newblock Learning optimal fair scoring systems for multi-class classification.
\newblock In \emph{2022 IEEE 34th International Conference on Tools with Artificial Intelligence (ICTAI)}, pages 197--204. IEEE, 2022.

\bibitem[Shah et~al.(2023)Shah, Shen, Ryu, Das, Sattigeri, Bu, and Wornell]{shah2023group}
Abhin Shah, Maohao Shen, Jongha~Jon Ryu, Subhro Das, Prasanna Sattigeri, Yuheng Bu, and Gregory~W Wornell.
\newblock Group fairness with uncertainty in sensitive attributes.
\newblock \emph{arXiv preprint arXiv:2302.08077}, 2023.

\bibitem[Wang et~al.(2020)Wang, Guo, Narasimhan, Cotter, Gupta, and Jordan]{wang2020robust}
Serena Wang, Wenshuo Guo, Harikrishna Narasimhan, Andrew Cotter, Maya Gupta, and Michael Jordan.
\newblock Robust optimization for fairness with noisy protected groups.
\newblock In \emph{Advances in Neural Information Processing Systems (NeurIPS)}, 2020.

\bibitem[{Wikipedia contributors}(2024)]{wikipedia2023fairness}
{Wikipedia contributors}.
\newblock Fairness (machine learning) --- {Wikipedia}{,} the free encyclopedia, 2024.
\newblock URL \url{https://en.wikipedia.org/w/index.php?title=Fairness_(machine_learning)&oldid=1201888401}.
\newblock [Online; accessed 1-February-2024].

\bibitem[Williamson and Menon(2019)]{williamson2019fairness}
Robert Williamson and Aditya Menon.
\newblock Fairness risk measures.
\newblock In \emph{International Conference on Machine Learning (ICML)}, 2019.

\bibitem[Xian and Zhao(2023)]{xian2023efficient}
Ruicheng Xian and Han Zhao.
\newblock Efficient post-processing for equal opportunity in fair multi-class classification, 2023.
\newblock URL \url{https://openreview.net/forum?id=zKjSmbYFZe}.

\bibitem[Yan et~al.(2020)Yan, Kao, and Ferrara]{yan2020fair}
Shen Yan, Hsien-te Kao, and Emilio Ferrara.
\newblock Fair class balancing: Enhancing model fairness without observing sensitive attributes.
\newblock In \emph{International Conference on Information and Knowledge Management (CIKM)}, 2020.

\bibitem[Yao et~al.(2023)Yao, Han, Zhou, Zhang, and Tsang]{Yao2023Latent}
Jiangchao Yao, Bo~Han, Zhihan Zhou, Ya~Zhang, and Ivor~W. Tsang.
\newblock Latent class-conditional noise model.
\newblock \emph{IEEE Trans. Pattern Anal. Mach. Intell.}, 45\penalty0 (8):\penalty0 9964--9980, 2023.

\bibitem[Yapo and Weiss(2018)]{yapo2018ethical}
Adrienne Yapo and Joseph~W. Weiss.
\newblock Ethical implications of bias in machine learning.
\newblock In \emph{Hawaii International Conference on System Sciences}, 2018.

\bibitem[Ye and Xie(2020)]{ye2020unbiased}
Qing Ye and Weijun Xie.
\newblock Unbiased subdata selection for fair classification: A unified framework and scalable algorithms.
\newblock \emph{arXiv preprint arXiv:2012.12356}, 2020.

\bibitem[Zhang and Long(2021)]{ZhaLon21}
Yiliang Zhang and Qi~Long.
\newblock Assessing fairness in the presence of missing data.
\newblock In \emph{Advances in Neural Information Processing Systems (NeurIPS)}, 2021.

\bibitem[Zhao et~al.(2022)Zhao, Dai, Shu, and Wang]{ZhaDaiShuWan22}
Tianxiang Zhao, Enyan Dai, Kai Shu, and Suhang Wang.
\newblock Towards fair classifiers without sensitive attributes: Exploring biases in related features.
\newblock In \emph{Proceedings of the Fifteenth ACM International Conference on Web Search and Data Mining}, 2022.

\bibitem[Zhu et~al.(2003)Zhu, Lafferty, and Ghahramani]{ZhuLafGha03}
X.~Zhu, J.~Lafferty, and Z.~Ghahramani.
\newblock Semi-supervised learning using gaussian fields and harmonic functions.
\newblock In \emph{Proc.\ Intl.\ Conf.\ Machine Learning}, 2003.

\bibitem[Zhu et~al.(2023)Zhu, Yao, Sun, Li, and Liu]{Zhuetal23}
Zhaowei Zhu, Yuanshun Yao, Jiankai Sun, Hang Li, and Yang Liu.
\newblock Weak proxies are sufficient and preferable for fairness with missing sensitive attributes.
\newblock \emph{arXiv:2210.03175}, 2023.

\end{thebibliography}


\begin{thebibliography}{}
\setlength{\itemindent}{-\leftmargin}
\makeatletter\renewcommand{\@biblabel}[1]{}\makeatother
\bibitem{} J.~Alspector, B.~Gupta, and R.~B.~Allen (1989).
    \newblock Performance of a stochastic learning microchip.
    \newblock In D. S. Touretzky (ed.),
    \textit{Advances in Neural Information Processing Systems 1}, 748--760.
    San Mateo, Calif.: Morgan Kaufmann.

\bibitem{} F.~Rosenblatt (1962).
    \newblock \textit{Principles of Neurodynamics.}
    \newblock Washington, D.C.: Spartan Books.

\bibitem{} G.~Tesauro (1989).
    \newblock Neurogammon wins computer Olympiad.
    \newblock \textit{Neural Computation} \textbf{1}(3):321--323.
\end{thebibliography}
